\documentclass[10pt,journal,compsoc]{IEEEtran}

\ifCLASSOPTIONcompsoc
  \usepackage[nocompress]{cite}
\else
  \usepackage{cite}
\fi
\ifCLASSINFOpdf
\else
\fi
\usepackage{amsmath}
\usepackage{amssymb}
\usepackage{array}
\usepackage{bbm}
\usepackage{calc}
\usepackage{caption}
\usepackage{color}
\usepackage{enumitem}
\usepackage{epsfig}
\usepackage{epstopdf}
\usepackage{graphicx}
\usepackage{hyperref}
\usepackage{multirow}
\usepackage{subfigure}
\usepackage{booktabs}
\usepackage{url}
\usepackage{xspace}

\newcommand{\figref}[1]{Fig\onedot~\ref{#1}}

\newcommand{\secref}[1]{Sec\onedot~\ref{#1}}
\newcommand{\tabref}[1]{Tab\onedot~\ref{#1}}

\newcommand{\by}[2]{\ensuremath{#1 \! \times \! #2}}

\def\onedot{\ifx\@let@token.\else.\null\fi\xspace}
\def\eg{\emph{e.g}\onedot} 
\def\ie{\emph{i.e}\onedot} 
 
 \def\vs{\emph{vs}\onedot}

\newcommand{\aatrous}{atrous~}

\begin{document}
%
\title{DeepLab: Semantic Image Segmentation with Deep Convolutional Nets, Atrous Convolution, and Fully Connected CRFs}
%
%
%
%

\author{Liang-Chieh~Chen, 
        George~Papandreou,~\IEEEmembership{Senior~Member,~IEEE,}
        Iasonas~Kokkinos,~\IEEEmembership{Member,~IEEE,}
        Kevin~Murphy,
        and Alan~L.~Yuille,~\IEEEmembership{Fellow,~IEEE}%
\IEEEcompsocitemizethanks{
\IEEEcompsocthanksitem L.-C. Chen, G. Papandreou, and K. Murphy are
with Google Inc. I. Kokkinos is with University College London. A. Yuille
is with the Departments of Cognitive Science and Computer Science, Johns Hopkins
University. The first two authors contributed equally to this work.}
}

\IEEEtitleabstractindextext{%
\begin{abstract}
In this work we  address the task of semantic image segmentation with Deep Learning
and make three main contributions that are experimentally shown to have substantial practical merit. 
\emph{First}, we highlight convolution with upsampled filters, or `\aatrous convolution', as a powerful
tool in dense prediction tasks. Atrous convolution allows us to explicitly control the resolution at
which feature responses are computed within Deep Convolutional Neural Networks. It also allows us 
to effectively enlarge the field of view of filters to incorporate larger context without increasing
the number of parameters or the amount of computation. 
\emph{Second}, we propose {\aatrous spatial pyramid pooling} (ASPP) to robustly segment objects at
multiple scales. ASPP probes an incoming convolutional feature layer with filters at multiple sampling
rates and effective fields-of-views, thus capturing objects as well as image context at multiple scales.
\emph{Third}, we improve the localization of object boundaries by combining methods from DCNNs and probabilistic
graphical models. The commonly deployed combination of max-pooling and downsampling in DCNNs achieves invariance but has a toll on localization 
accuracy.  We overcome this by combining the responses at the final DCNN layer with a
fully connected Conditional Random Field (CRF), which is shown both qualitatively and quantitatively to improve localization performance.
Our proposed ``DeepLab'' system sets the new state-of-art at the PASCAL
  VOC-2012 semantic image segmentation task, reaching 79.7\% mIOU in the
  test set, and advances the results on three other datasets:
  PASCAL-Context, PASCAL-Person-Part, and Cityscapes. All of our code is made publicly available online.
\end{abstract}

\begin{IEEEkeywords}
Convolutional Neural Networks, Semantic Segmentation, Atrous Convolution, Conditional Random Fields.
\end{IEEEkeywords}}

\maketitle

\IEEEdisplaynontitleabstractindextext

%
\IEEEpeerreviewmaketitle

\IEEEraisesectionheading{\section{Introduction}\label{sec:introduction}}

Deep Convolutional Neural Networks (DCNNs) \cite{LeCun1998} have 
pushed the performance of computer vision systems to soaring heights on a
broad array of high-level problems, including image classification
\cite{KrizhevskyNIPS2013, sermanet2013overfeat, simonyan2014very,
  szegedy2014going, papandreou2014untangling} and object detection
\cite{girshick2014rcnn, erhan2014scalable, girshick2015fast, ren2015faster,
  he2015deep, liu2015ssd}, where DCNNs trained
in an end-to-end manner have delivered strikingly better results than systems relying
on hand-crafted features.
Essential to this success is the built-in invariance of DCNNs
to local image transformations, which allows them to learn increasingly
 abstract data representations \cite{zeiler2014visualizing}. This
invariance is clearly desirable for classification tasks, but can hamper
dense prediction tasks such as semantic segmentation, where abstraction of spatial
information is undesired.

In particular we consider three challenges in the application of DCNNs to semantic image segmentation: (1) reduced feature resolution,
(2) existence of objects at multiple scales, and (3) reduced localization
accuracy due to DCNN invariance. Next, we discuss these challenges and our
approach to overcome them in our proposed DeepLab system.

The first challenge is caused by the repeated combination of max-pooling and downsampling (`striding')
 performed at consecutive layers of  DCNNs originally designed
for image classification \cite{KrizhevskyNIPS2013, simonyan2014very, szegedy2014going}.
This results in feature maps with significantly reduced spatial resolution when
the DCNN is employed in a fully convolutional fashion \cite{long2014fully}. In
order to overcome this hurdle and efficiently produce denser feature maps, we
remove the downsampling operator from the last few max pooling layers of DCNNs
and instead {\it upsample the filters} in subsequent convolutional layers, resulting
in feature maps computed at a higher sampling rate. Filter upsampling amounts to
inserting holes (`trous' in French) between nonzero filter taps. This technique has a long history in signal processing,
originally developed for the efficient computation of the undecimated wavelet
transform in a scheme also known as ``algorithme \`a trous''
\cite{holschneider1989real}. We use the term \emph{atrous convolution} as a shorthand for convolution with
upsampled filters. 
Various flavors of this idea have been used before
in the context of DCNNs by \cite{giusti2013fast, sermanet2013overfeat,
papandreou2014untangling}. In practice, we recover full resolution feature maps
by a combination of atrous convolution, which computes feature maps more
densely, followed by simple bilinear interpolation of the feature responses to
the original image size. This scheme offers a simple yet powerful alternative to
using deconvolutional layers \cite{zeiler2014visualizing, long2014fully} in
dense prediction tasks. Compared to regular convolution with larger filters, atrous
convolution allows us to effectively enlarge the field of view of filters without
increasing the number of parameters or the amount of computation.

The second challenge is caused by the existence of objects at
multiple scales. A standard way to deal with this is to present to the DCNN
rescaled versions of the same image and then aggregate the feature or score maps
\cite{papandreou2014untangling, chen2015attention,kokkinos2016pushing}. We show that this approach
indeed increases the performance of our system, but comes at the cost of computing feature
responses at all DCNN layers for multiple scaled versions of the input image. Instead, motivated
by  spatial pyramid pooling \cite{lazebnik2006beyond, he2014spatial},
we propose a computationally efficient scheme of resampling a given feature layer at multiple
rates prior to  convolution. This amounts to probing the original image with multiple filters
that have complementary effective fields of view, thus capturing objects as
well as useful image context at multiple scales. Rather than actually resampling features, 
we efficiently implement this mapping using
multiple parallel atrous convolutional layers with different sampling
rates; we call the proposed technique ``atrous spatial pyramid pooling'' (ASPP).

The third challenge relates to the fact that an object-centric classifier 
requires invariance to spatial transformations, inherently
limiting the spatial accuracy of a DCNN. One way to mitigate this
problem is to use skip-layers to extract ``hyper-column'' features from multiple network layers when
computing the final segmentation result
\cite{hariharan2014hypercolumns, long2014fully}. Our work explores an
alternative approach which we show to be highly effective. In particular, we
boost our model's ability to capture fine details by employing a fully-connected
Conditional Random Field (CRF) \cite{krahenbuhl2011efficient}. CRFs have been
broadly used in semantic segmentation to combine class scores computed by
multi-way classifiers with the low-level information captured by the local
interactions of pixels and edges
\cite{rother2004grabcut, shotton2009textonboost} or superpixels
\cite{lucchi2011spatial}. Even though works of increased sophistication have
been proposed to model the hierarchical dependency \cite{he2004multiscale,
  ladicky2009associative, lempitsky2011pylon} and/or high-order dependencies
of segments \cite{delong2012fast, gonfaus2010harmony, kohli2009robust, CPY13, Wang15}, we
use the fully connected pairwise CRF proposed by
\cite{krahenbuhl2011efficient} for its efficient computation, and ability to
capture fine edge details while also catering for long range dependencies. That
model was shown in \cite{krahenbuhl2011efficient} to improve the performance of
a boosting-based pixel-level classifier. In this work, we demonstrate that it
leads to state-of-the-art results when coupled with a DCNN-based pixel-level
classifier.

\begin{figure*}[!th]
  \centering
  \includegraphics[width=0.7\linewidth]{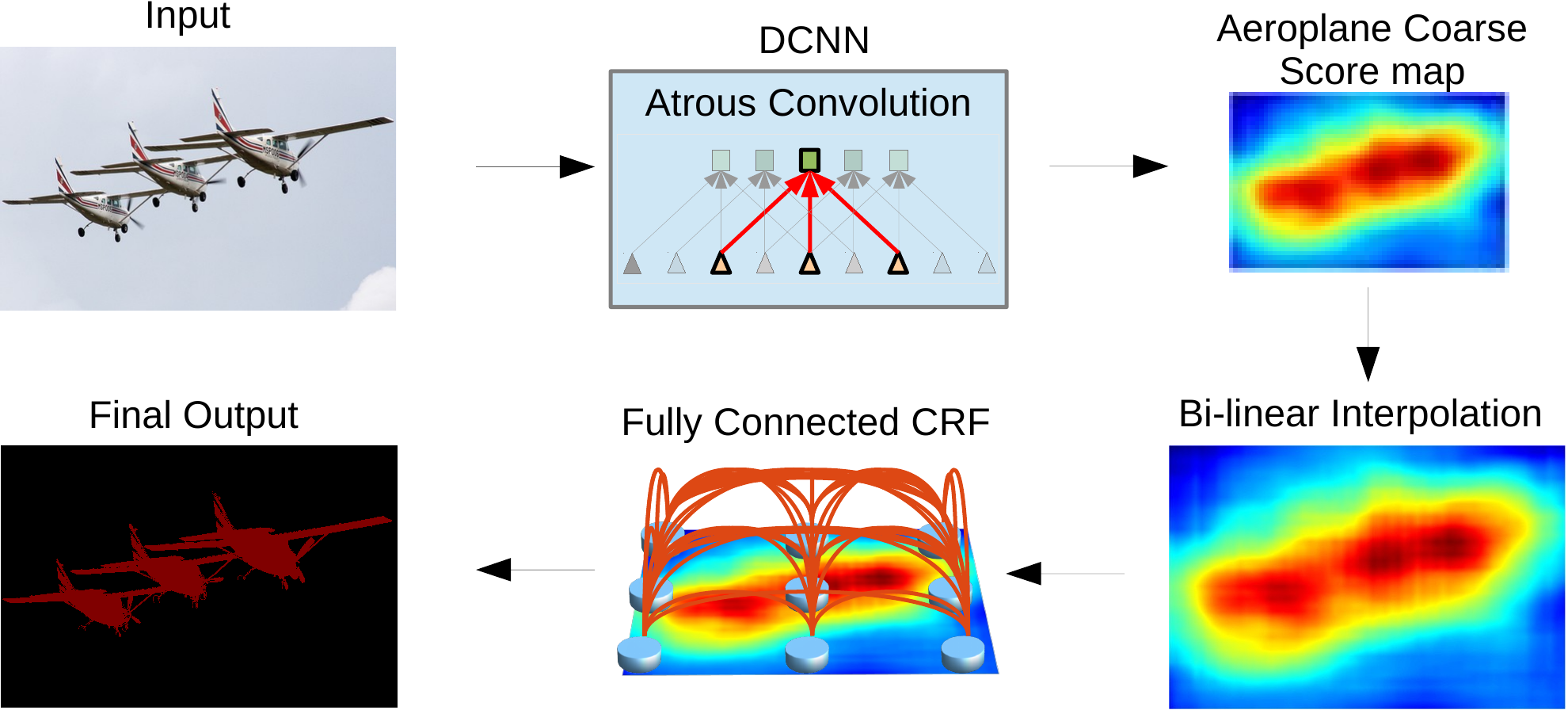}
  \caption{Model Illustration. A Deep Convolutional Neural Network such as
  VGG-16 or ResNet-101 is employed in a fully convolutional fashion, using
  atrous convolution to reduce the degree of signal downsampling (from 32x
  down 8x). A bilinear interpolation stage enlarges the feature maps to the
  original image resolution. A fully connected CRF is then applied to refine
  the segmentation result and better capture the object boundaries.}
  \label{fig:ModelIllustration}
\end{figure*}

A high-level illustration of the proposed DeepLab model is shown in
\figref{fig:ModelIllustration}. A deep convolutional neural network
(VGG-16 \cite{simonyan2014very} or ResNet-101 \cite{he2015deep} in this work)
trained in the task of image classification is re-purposed to the task of
semantic segmentation by (1) transforming all the fully connected layers
to convolutional layers (\ie, fully convolutional network \cite{long2014fully})
and (2) increasing feature resolution through
atrous convolutional layers, allowing us  to compute feature
responses  every 8 pixels instead of every 32 pixels in the original network. 
We then employ bi-linear interpolation to upsample by a
factor of 8 the score map to reach the original image resolution, yielding the 
input to 
a fully-connected CRF \cite{krahenbuhl2011efficient} that refines the
segmentation results.

From a practical standpoint, the three main advantages of our DeepLab
system are: (1) Speed: by virtue of atrous convolution, our dense DCNN operates
at 8 FPS on an NVidia Titan X GPU, while Mean Field Inference for the
fully-connected CRF requires 0.5 secs on a CPU. (2) Accuracy: we obtain
state-of-art results on several challenging datasets, including the PASCAL VOC 2012
semantic segmentation benchmark \cite{everingham2014pascal}, PASCAL-Context
\cite{mottaghi2014role}, PASCAL-Person-Part \cite{chen_cvpr14}, and Cityscapes
\cite{Cordts2016Cityscapes}. (3) Simplicity: our system is composed of a cascade of
two very well-established modules, DCNNs and CRFs.

The updated DeepLab system we present in this paper features several
improvements compared to its first version reported in our original conference
publication \cite{chen2014semantic}. Our new version can better segment objects
at multiple scales, via either multi-scale input processing \cite{farabet2013learning, lin2015efficient, chen2015attention} or the 
proposed ASPP. We have built a residual net variant of DeepLab by adapting
the state-of-art ResNet \cite{he2015deep} image classification DCNN, achieving
better semantic segmentation performance compared to our original model based on
VGG-16 \cite{simonyan2014very}. Finally, we present a more comprehensive
experimental evaluation of multiple model variants and report state-of-art
results not only on the PASCAL VOC 2012 benchmark but also on other challenging
tasks. We have implemented the proposed methods by extending
the Caffe framework \cite{jia2014caffe}. We share our code and models at a
companion web site \url{http://liangchiehchen.com/projects/DeepLab.html}.



\section{Related Work}
Most of the successful semantic segmentation systems developed in the previous decade relied on hand-crafted features 
combined with flat classifiers, such as Boosting \cite{TuB10,shotton2009textonboost}, Random Forests \cite{shotton2008semantic}, or Support Vector Machines \cite{FulkersonVS09}. Substantial improvements have been achieved
 by incorporating richer information from context
\cite{carreira2012semantic} 
and structured prediction techniques \cite{he2004multiscale,ladicky2009associative,carreira2012cpmc,krahenbuhl2011efficient}, but
the performance of these systems has always been compromised by the limited expressive power of the features.
Over the past few years the breakthroughs of Deep Learning in image classification were quickly transferred to the semantic 
segmentation task. Since this task involves both segmentation and classification, a central question is how to combine the two tasks. 

The first family of DCNN-based systems for semantic segmentation typically employs a cascade of
bottom-up image segmentation, followed by DCNN-based region classification.
For instance the bounding box proposals and masked regions delivered by
\cite{arbelaez2014multiscale, Uijlings13} are used in \cite{girshick2014rcnn}
and \cite{hariharan2014simultaneous}  as inputs to a DCNN to incorporate shape
information into the classification process. Similarly, the authors of
 \cite{mostajabi2014feedforward} rely on a superpixel representation.
Even though these approaches can benefit from the sharp boundaries delivered by a  good segmentation,
they also cannot recover from any of its errors.

The second family of works relies on using convolutionally computed DCNN
features for dense image labeling, and couples them with segmentations that are obtained independently.
Among the first have been
\cite{farabet2013learning} who apply DCNNs at multiple image resolutions and
then employ a segmentation tree to smooth the prediction results. More recently,
\cite{hariharan2014hypercolumns} propose to use skip layers and concatenate the computed
intermediate feature maps within the DCNNs for pixel classification. Further, \cite{dai2014convolutional}
propose to pool the intermediate feature maps by region proposals. These works
still employ segmentation algorithms that are decoupled from the DCNN
classifier's results, thus risking commitment to premature decisions. 

The third family of works uses DCNNs to directly provide dense category-level pixel labels, which makes it 
possible to even discard segmentation altogether. 
The segmentation-free approaches of
\cite{long2014fully, eigen2014predicting} directly apply DCNNs to the whole
image in a fully convolutional fashion, transforming the last fully connected
layers of the DCNN into convolutional layers. In order to deal with the spatial
localization issues outlined in the introduction, \cite{long2014fully} upsample
and concatenate the scores from intermediate feature maps, while
\cite{eigen2014predicting} refine the prediction result from coarse to fine by
propagating the coarse results to another DCNN.
Our work builds on  these works, and as described in the introduction 
extends them by exerting control on the feature resolution, introducing multi-scale pooling techniques
and integrating  the
densely connected CRF of \cite{krahenbuhl2011efficient} on top of the DCNN. We
show that this leads to significantly better segmentation results, especially
along object boundaries. The combination of DCNN and CRF is of course not new
but previous works only tried locally connected CRF models. Specifically,
\cite{cogswell2014combining} use CRFs as a proposal mechanism for a DCNN-based
reranking system, while \cite{farabet2013learning} treat superpixels as nodes
for a local pairwise CRF and use graph-cuts for discrete inference. As such
their models were limited by errors in superpixel computations or ignored
long-range dependencies. Our approach instead treats every pixel as a CRF node
receiving unary potentials by the DCNN. Crucially, the Gaussian CRF potentials
in the fully connected CRF model of \cite{krahenbuhl2011efficient} that we adopt
can capture long-range dependencies and at the same time the model is amenable
to fast mean field inference. We note that mean field inference had been
extensively studied for traditional image segmentation tasks
\cite{geiger1991parallel, geiger1991common, kokkinos2008computational}, but
these older models were typically limited to short-range connections. In
independent work, \cite{bell2014material} use a very similar densely connected
CRF model to refine the results of DCNN for the problem of material
classification. However, the DCNN module of \cite{bell2014material} was only
trained by sparse point supervision instead of dense supervision at every pixel.

Since the first version of this work was made publicly available
\cite{chen2014semantic}, the area of semantic segmentation has progressed
drastically. Multiple groups have made important advances, significantly raising
the bar on the PASCAL VOC 2012 semantic segmentation benchmark, as reflected to
the high level of activity in the benchmark's
leaderboard\footnote{\url{http://host.robots.ox.ac.uk:8080/leaderboard/displaylb.php?challengeid=11&compid=6}}
\cite{papandreou2015weakly, zheng2015conditional, dai2015boxsup, noh2015learning,
liu2015semantic, lin2015efficient, chen2015attention, chen2015semantic}.
Interestingly, most top-performing methods have adopted one or both of the key
ingredients of our DeepLab system: Atrous convolution for efficient dense
feature extraction and refinement of the raw DCNN scores by means of a fully
connected CRF. We outline below some of the most important and interesting advances.

\newcommand{\mycomment}[1]{}
\mycomment{
	A key difference compared to \cite{long2014fully} lies in the way we produce
	feature maps at the original image resolution: They use a sequence of
	deconvolutional layers \cite{zeiler2014visualizing}, while we use a combination
	of atrous convolution and bilinear interpolation, resulting in a significantly
	simpler system that explicitly handles the signal downsampling issue, requires
	fewer parameters, and is easier to train.
}

\emph{End-to-end training for structured prediction} has  more recently 
been explored in several related works. 
While we employ the CRF as a post-processing method,
\cite{zheng2015conditional, chen2014learning, schwing2015fully, liu2015semantic,
lin2015efficient} have successfully pursued joint learning of the DCNN and CRF.
In particular, \cite{zheng2015conditional, schwing2015fully} unroll the CRF
mean-field inference steps to convert the whole system into an end-to-end
trainable feed-forward network, while \cite{liu2015semantic} approximates one
iteration of the dense CRF mean field inference \cite{krahenbuhl2011efficient}
by convolutional layers with learnable filters. Another fruitful direction
pursued by \cite{lin2015efficient,chandra2016fast} is to learn the pairwise terms of a CRF
via a DCNN, significantly improving performance at the cost of heavier
computation. In a different direction, \cite{chen2015semantic} replace the
bilateral filtering module used in mean field inference with a faster
domain transform module \cite{GastalOliveira2011DomainTransform}, improving the speed and lowering the
memory requirements of the overall system, while
\cite{bertasius2015high, kokkinos2016pushing} combine semantic segmentation
with edge detection.

\emph{Weaker supervision} has been pursued in a number of papers, relaxing
the assumption that pixel-level semantic annotations are available for the whole
training set \cite{pinheiro2014weakly, papandreou2015weakly, pathakICCV15ccnn, hong2015decoupled},
achieving significantly better results than weakly-supervised pre-DCNN systems
such as \cite{vezhnevets2011weakly}. In another line of research,
\cite{hariharan2014simultaneous, liang2015proposal} pursue instance segmentation, jointly tackling  object detection and semantic
segmentation.

What we call here \emph{atrous convolution} was originally developed for the efficient
computation of the undecimated wavelet transform in the ``algorithme \`a trous''
scheme of \cite{holschneider1989real}. We refer the interested reader to
\cite{fowler2005redundant} for early references from the wavelet literature.
Atrous convolution is also intimately related to the ``noble identities'' in
multi-rate signal processing, which builds on the same interplay of input signal
and filter sampling rates \cite{vaidyanathan1990multirate}. Atrous convolution
is a term we first used in \cite{papandreou2014untangling}. The same operation
was later called dilated convolution by \cite{yu2015multi}, a term they coined
motivated by the fact that the operation corresponds to regular convolution
with upsampled (or dilated in the terminology of \cite{holschneider1989real})
filters. Various authors have used the same operation before for denser feature
extraction in DCNNs \cite{giusti2013fast, sermanet2013overfeat, papandreou2014untangling}.
Beyond mere resolution enhancement, atrous convolution allows us to enlarge the
field of view of filters to incorporate larger context, which we have shown in
\cite{chen2014semantic} to be beneficial. This approach has been pursued further
by \cite{yu2015multi}, who employ a series of atrous convolutional layers with
increasing rates to aggregate multiscale context. The atrous spatial pyramid
pooling scheme proposed here to capture multiscale objects and context also
employs multiple atrous convolutional layers with different sampling rates,
which we however lay out in parallel instead of in serial.
Interestingly, the atrous convolution technique has also been adopted for a broader set of  tasks, such as object detection
 \cite{liu2015ssd,dai2016rfcn}, instance-level segmentation \cite{dai2016instance}, visual question answering \cite{chen2015abc}, and optical flow
 \cite{sevilla2016optical}.

We also show that, as expected, integrating into DeepLab more advanced image
classification DCNNs such as the residual net of \cite{he2015deep} leads to
better results. This has also been observed independently by \cite{wu2016high}.

\section{Methods}
\label{sec:methods}

\subsection{Atrous Convolution for Dense Feature Extraction and Field-of-View Enlargement}
\label{sec:convnet-hole}
The use of DCNNs for semantic segmentation, or other dense prediction tasks, has been shown to be
simply and successfully addressed by deploying DCNNs in a fully convolutional fashion \cite{sermanet2013overfeat, long2014fully}.
 However, the repeated combination of max-pooling and striding
 at consecutive layers of these networks reduces significantly the spatial resolution of the
 resulting feature maps, typically by a factor of 32 across each direction in recent DCNNs.
 A partial remedy is to use `deconvolutional' layers as  in  \cite{long2014fully},
which however requires additional memory and time.
 
We advocate instead the use of atrous convolution, originally developed for the efficient computation of the
undecimated wavelet transform in the ``algorithme \`a trous'' scheme of
\cite{holschneider1989real} and used before in the DCNN context by
\cite{giusti2013fast, sermanet2013overfeat, papandreou2014untangling}.
This algorithm allows us to compute the responses of any layer at any desirable resolution.
It can be applied post-hoc, once a network has been trained, but can also be seamlessly integrated with training.

\begin{figure}
  \begin{tabular}{c}
    \includegraphics[width=0.9\linewidth]{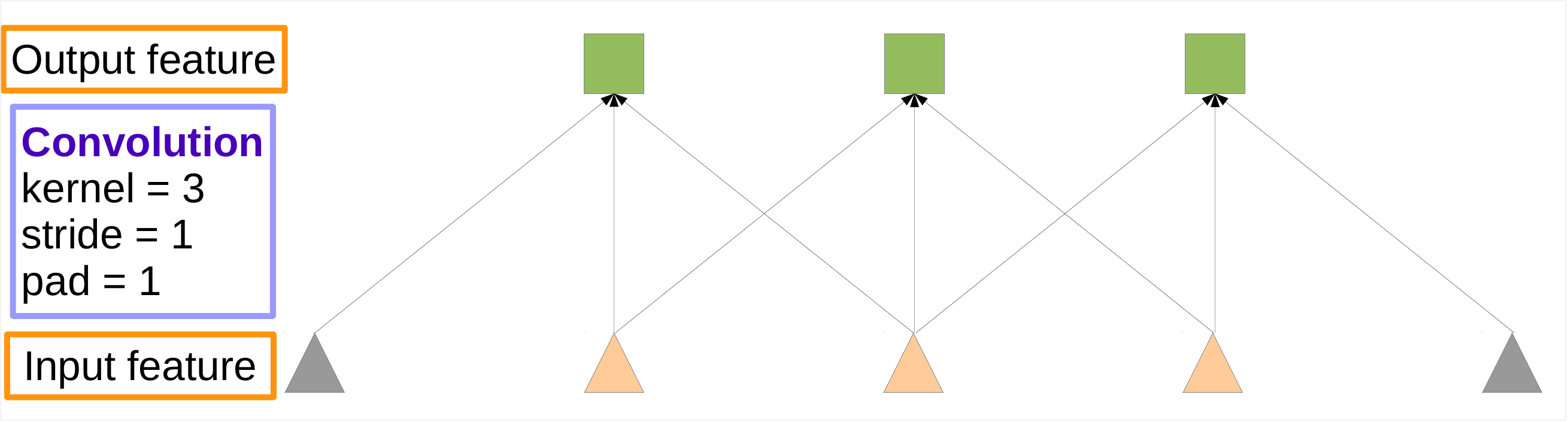} \\
    {\scriptsize (a) Sparse feature extraction} \\
    \includegraphics[width=0.9\linewidth]{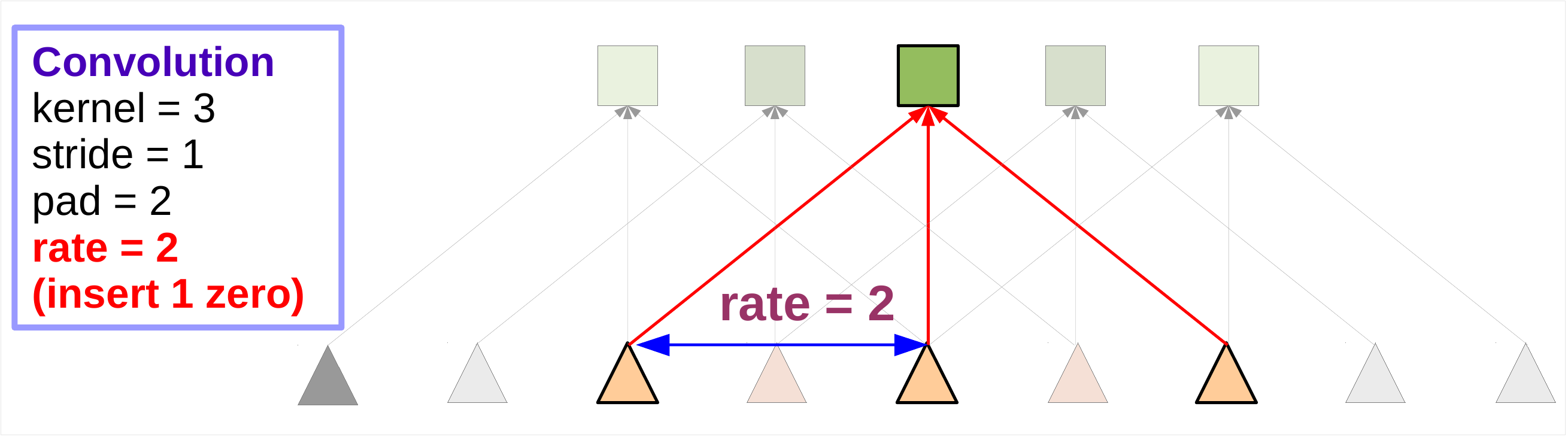} \\
    {\scriptsize (b) Dense feature extraction} \\
  \end{tabular}
  \caption{Illustration of atrous convolution in 1-D. (a) Sparse feature
    extraction with standard convolution on a low resolution input feature map.
    (b) Dense feature extraction with atrous convolution with rate $r = 2$,
    applied on a high resolution input feature map.}
  \label{fig:hole}
\end{figure}

Considering one-dimensional signals first, the output $y[i]$ of atrous convolution \footnote{We follow the
standard practice in the DCNN literature and use non-mirrored filters in this
definition.} of a 1-D input signal $x[i]$ with a filter $w[k]$ of length $K$ is
defined as:
\begin{equation}
  y[i] = \sum_{k=1}^K x[i + r \cdot k] w[k].
\end{equation}
The \emph{rate} parameter $r$ corresponds to the stride with which we sample the
input signal. Standard convolution is a special case for rate $r = 1$.
See \figref{fig:hole} for illustration.

\begin{figure}
  \centering
    \begin{tabular}{c}
    	\includegraphics[width=.99 \linewidth]{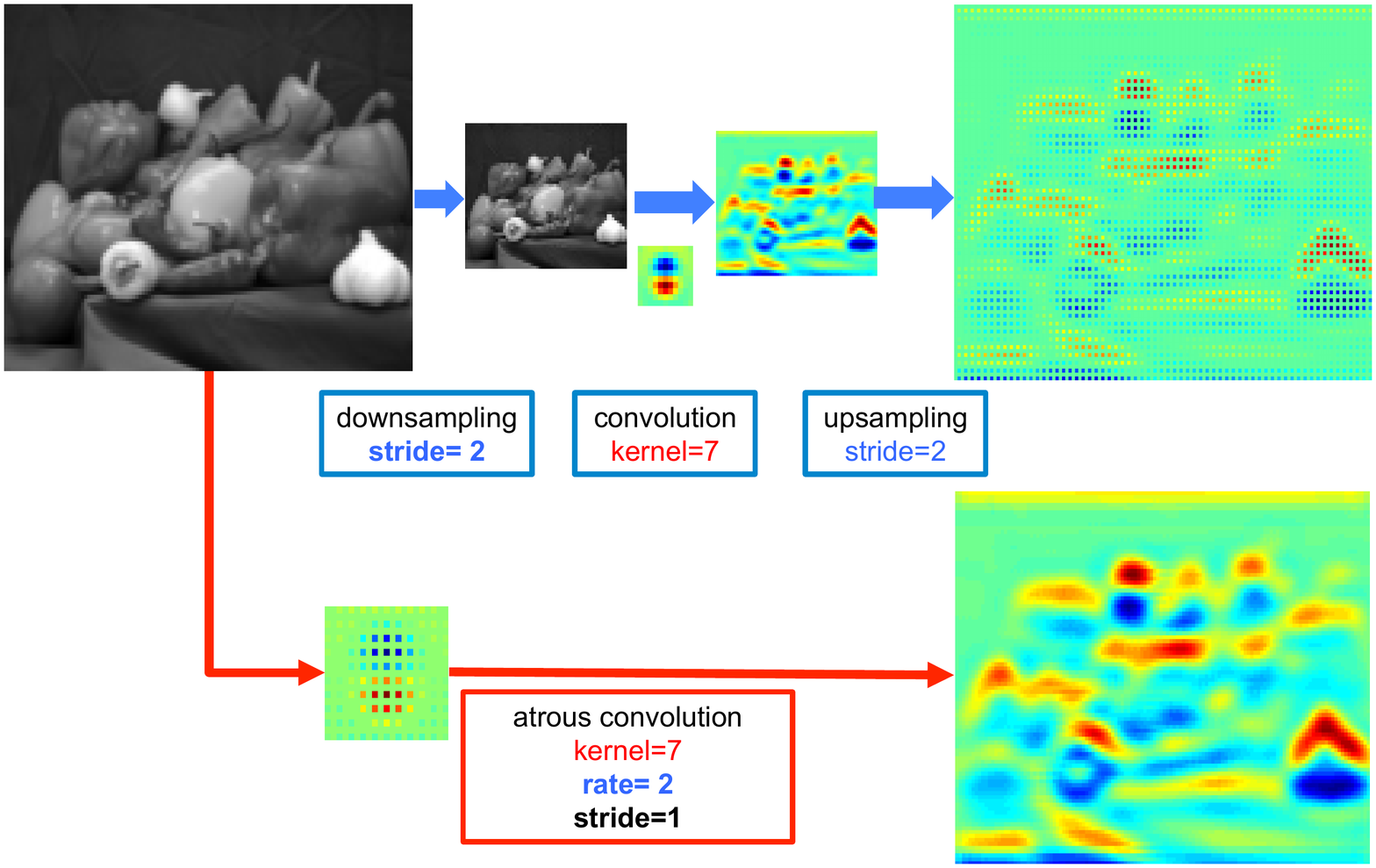}
    	\vspace{-.9cm}
    \end{tabular}
   \caption{Illustration of atrous convolution in 2-D. Top row: sparse feature
   	extraction with standard convolution on a low resolution input feature map.
   	Bottom row: Dense feature extraction with atrous convolution with rate $r = 2$,
   	applied on a high resolution input feature map.}
    \label{fig:hole2d}
   \end{figure}

We illustrate the algorithm's operation in 2-D through a simple example in \figref{fig:hole2d}: Given an image, we 
assume that we first have a downsampling operation that reduces the resolution by a factor of 2, and then perform  a 
convolution with a  kernel - here, the vertical Gaussian derivative. If one  implants the resulting 
feature map in the original image coordinates, we realize that we have obtained responses at only 1/4 of the image positions. 
Instead, we can compute responses at all image positions 
if we convolve the full resolution image with a filter `with holes', in which 
we upsample the original filter by a
factor of 2, and introduce zeros  in between filter values. 
Although the effective filter size increases, we only need to take into account the
non-zero filter values, hence both the number of filter parameters and the number of operations per position stay constant. 
The resulting scheme allows us to easily and explicitly control the spatial resolution of neural
network feature responses.

In the context of DCNNs one can use  atrous convolution in a chain of layers,
 effectively allowing us to compute the final DCNN network responses at an
arbitrarily high resolution. For example, in order to double the spatial density
of computed feature responses in the VGG-16 or ResNet-101 networks, we find the
last pooling or convolutional layer that decreases resolution ('pool5' or 'conv5\_1'
respectively), set its stride to 1 to avoid signal decimation, and replace all
subsequent convolutional layers with atrous convolutional layers having rate
$r = 2$. 
Pushing this approach all the
way through the network could allow us to compute feature responses at the original image
resolution, but this ends up being too
costly. We have adopted instead a hybrid approach that strikes a
good efficiency/accuracy trade-off, using atrous convolution to increase by a
factor of 4 the density of computed feature maps, followed by fast bilinear
interpolation by an additional factor of 8 to
recover feature maps at the original image resolution. Bilinear interpolation
is sufficient in this setting because the class score maps (corresponding to
log-probabilities) are quite smooth, as illustrated in
\figref{fig:score-maps}. Unlike the deconvolutional approach adopted by
\cite{long2014fully}, the proposed approach converts image classification
networks into dense feature extractors without requiring learning any extra
parameters, leading to faster DCNN training in practice. 

Atrous convolution also allows us to arbitrarily enlarge the \emph{field-of-view} of
filters at any DCNN layer.
State-of-the-art DCNNs typically employ spatially small
convolution kernels (typically \by{3}{3}) in order to keep both computation and
number of parameters contained. Atrous convolution with rate $r$ introduces
$r-1$ zeros between consecutive filter values, effectively enlarging the kernel
size of a \by{k}{k} filter to $k_e = k + (k-1)(r-1)$ without increasing the
number of parameters or the amount of computation. It thus offers an efficient
mechanism to control the field-of-view and finds the best trade-off between
accurate localization (small field-of-view) and context assimilation (large
field-of-view). We have successfully experimented with this technique:
Our DeepLab-LargeFOV model variant \cite{chen2014semantic} employs atrous
convolution with rate $r = 12$ in VGG-16 `fc6' layer with significant
performance gains, as detailed in Section~\ref{sec:experiments}.

Turning to implementation aspects, 
there are two efficient ways to  perform atrous convolution. The first
is to implicitly upsample the filters by inserting holes (zeros), or
equivalently sparsely sample the input feature maps \cite{holschneider1989real}.
We implemented this in our earlier work \cite{papandreou2014untangling,
 chen2014semantic}, followed by \cite{yu2015multi}, within the Caffe framework
\cite{jia2014caffe} by adding to the \textsl{im2col} function (it extracts
vectorized patches from multi-channel feature maps) the option to sparsely
sample the underlying feature maps. The second method, originally proposed by
\cite{shensa1992discrete} and used in \cite{giusti2013fast, sermanet2013overfeat}
is to subsample the input feature map by a factor equal to the atrous
convolution rate $r$, deinterlacing it to produce $r^2$ reduced resolution maps,
one for each of the $\by{r}{r}$ possible shifts. This is followed by applying
standard convolution to these intermediate feature maps and reinterlacing them
to the original image resolution. By reducing atrous convolution into regular
convolution, it allows us to use off-the-shelf highly optimized convolution
routines. We have implemented the second approach into the TensorFlow framework
\cite{abadi2016tensorflow}.


\subsection{Multiscale Image Representations using Atrous Spatial Pyramid Pooling}

DCNNs have shown a remarkable
ability to implicitly represent scale, simply by being trained on datasets that
contain objects of varying size. Still, explicitly accounting for object scale
can improve the DCNN's ability to successfully handle both
large and small objects \cite{papandreou2014untangling}.

We have experimented with two approaches to handling
 scale variability in semantic segmentation.
The first approach amounts to standard multiscale
processing \cite{chen2015attention, kokkinos2016pushing}. We extract DCNN score
maps from multiple (three in our experiments) rescaled versions of the original
image using parallel DCNN branches that share the same parameters. To produce
the final result, we bilinearly interpolate the feature maps from the parallel
DCNN branches to the original image resolution and fuse them, by taking at each
position the maximum response across the different scales. We do this both
during training and testing. Multiscale processing significantly improves
performance, but at the cost of computing feature responses at all DCNN layers for
multiple scales of input. 

\begin{figure}[!t]
  \centering
  \scalebox{1}{
  \begin{tabular}{c}
    \includegraphics[height=4.5cm]{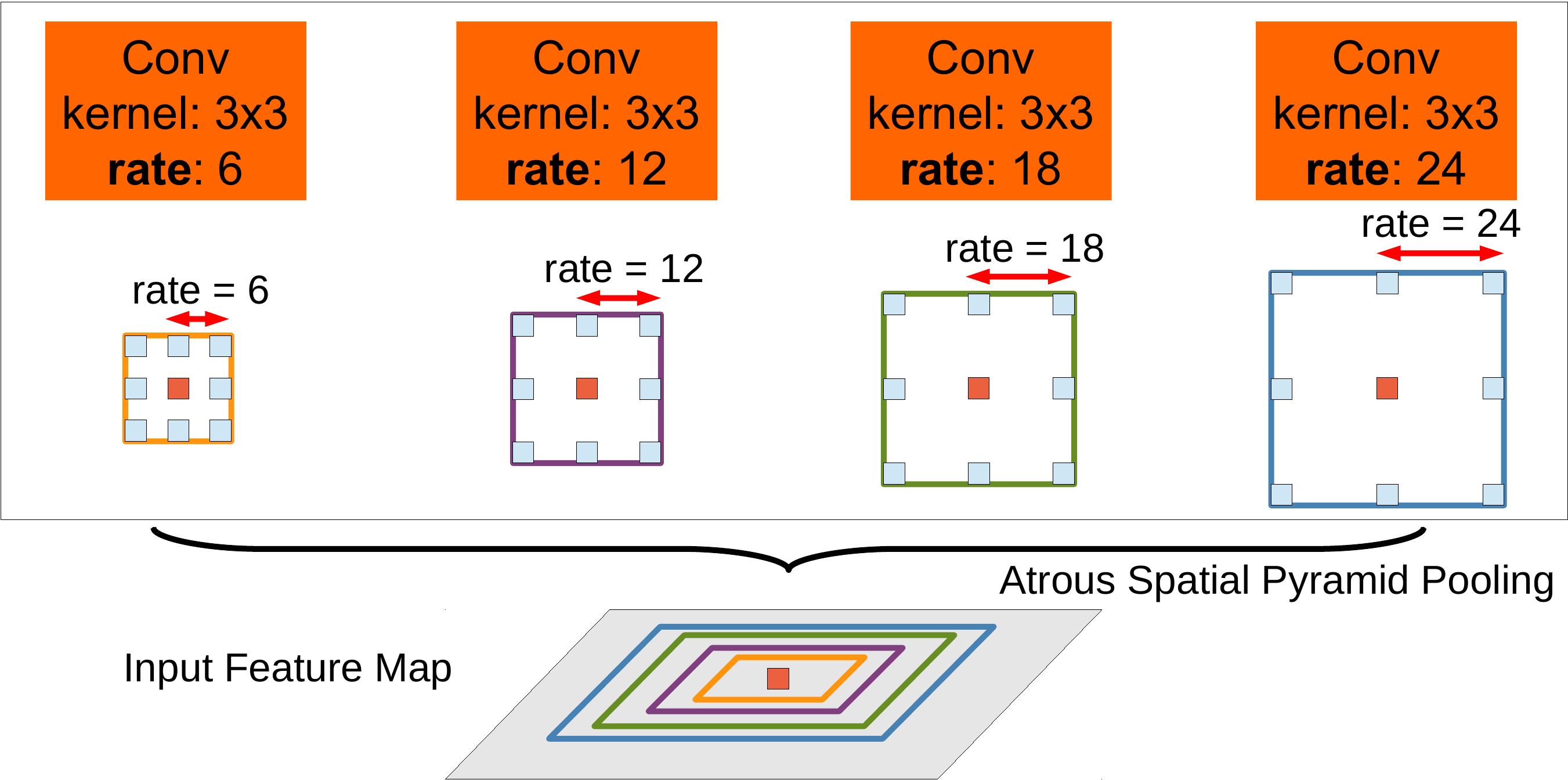} \\
  \end{tabular}
  }
  \caption{Atrous Spatial Pyramid Pooling (ASPP). To classify the center pixel (orange),
    ASPP exploits multi-scale features by employing multiple parallel filters with
    different rates. The effective Field-Of-Views are shown in different colors.}
  \label{fig:aspp_fov}
\end{figure}

The second approach is inspired by the success of the R-CNN spatial pyramid pooling method of \cite{he2014spatial},
which showed that regions of an arbitrary scale can be accurately and efficiently classified by resampling 
convolutional features extracted at a single scale.
 We have implemented a variant of their scheme which uses multiple
parallel atrous convolutional layers with different sampling rates. The features extracted for each sampling
rate are further processed in separate branches and fused to generate the final
result. The proposed ``atrous spatial pyramid pooling'' (DeepLab-ASPP) approach
generalizes our DeepLab-LargeFOV variant and is illustrated in \figref{fig:aspp_fov}.


\subsection{Structured Prediction with Fully-Connected Conditional Random Fields for Accurate Boundary Recovery}
\label{sec:boundary-recovery}

\begin{figure}[t]
  \centering
  \begin{tabular}{c c c c c}
    \includegraphics[width=0.16\linewidth]{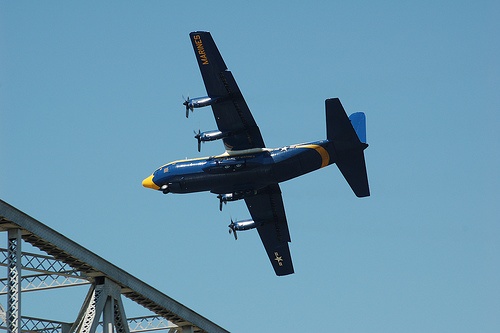} & 
    \includegraphics[width=0.16\linewidth]{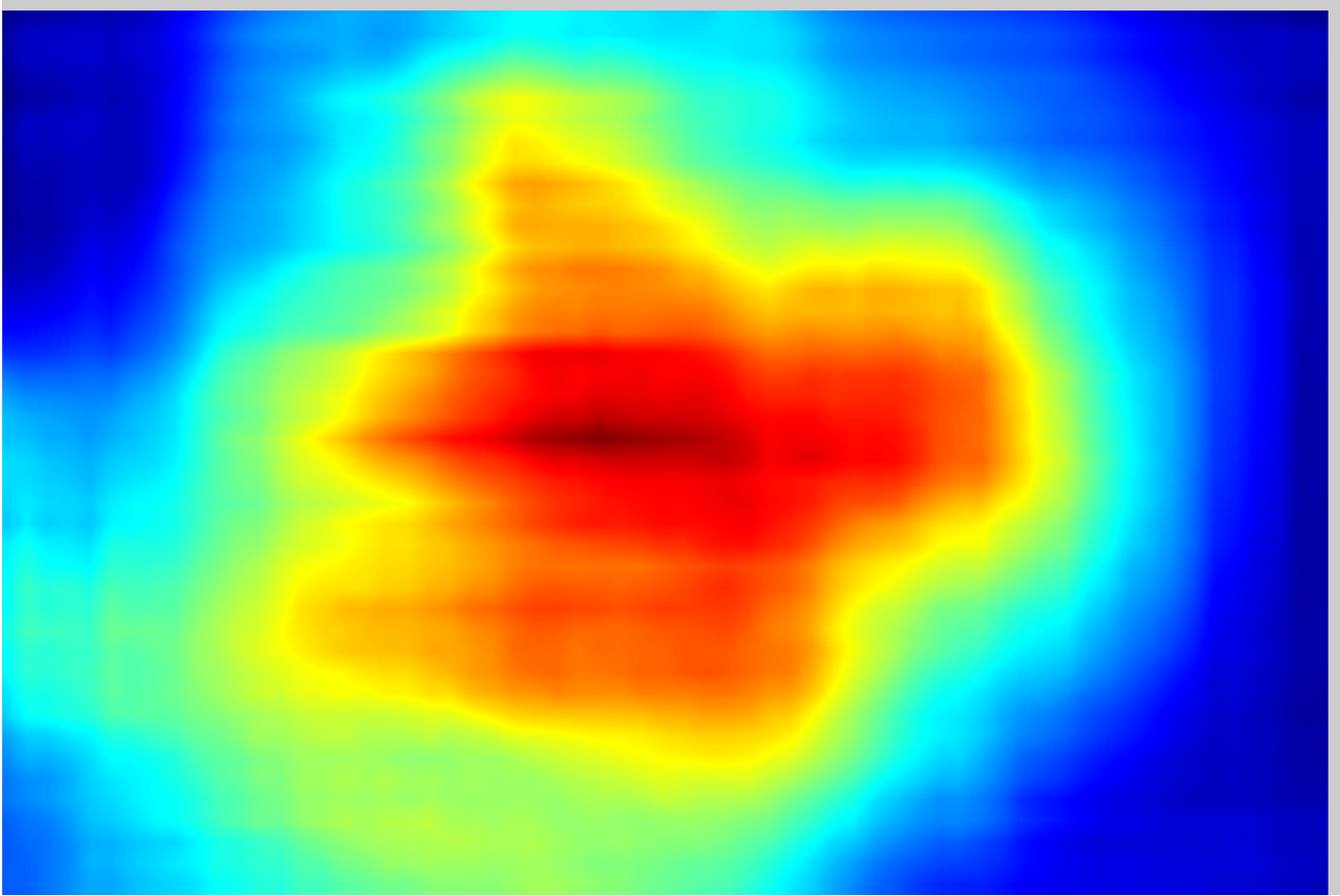} &
    \includegraphics[width=0.16\linewidth]{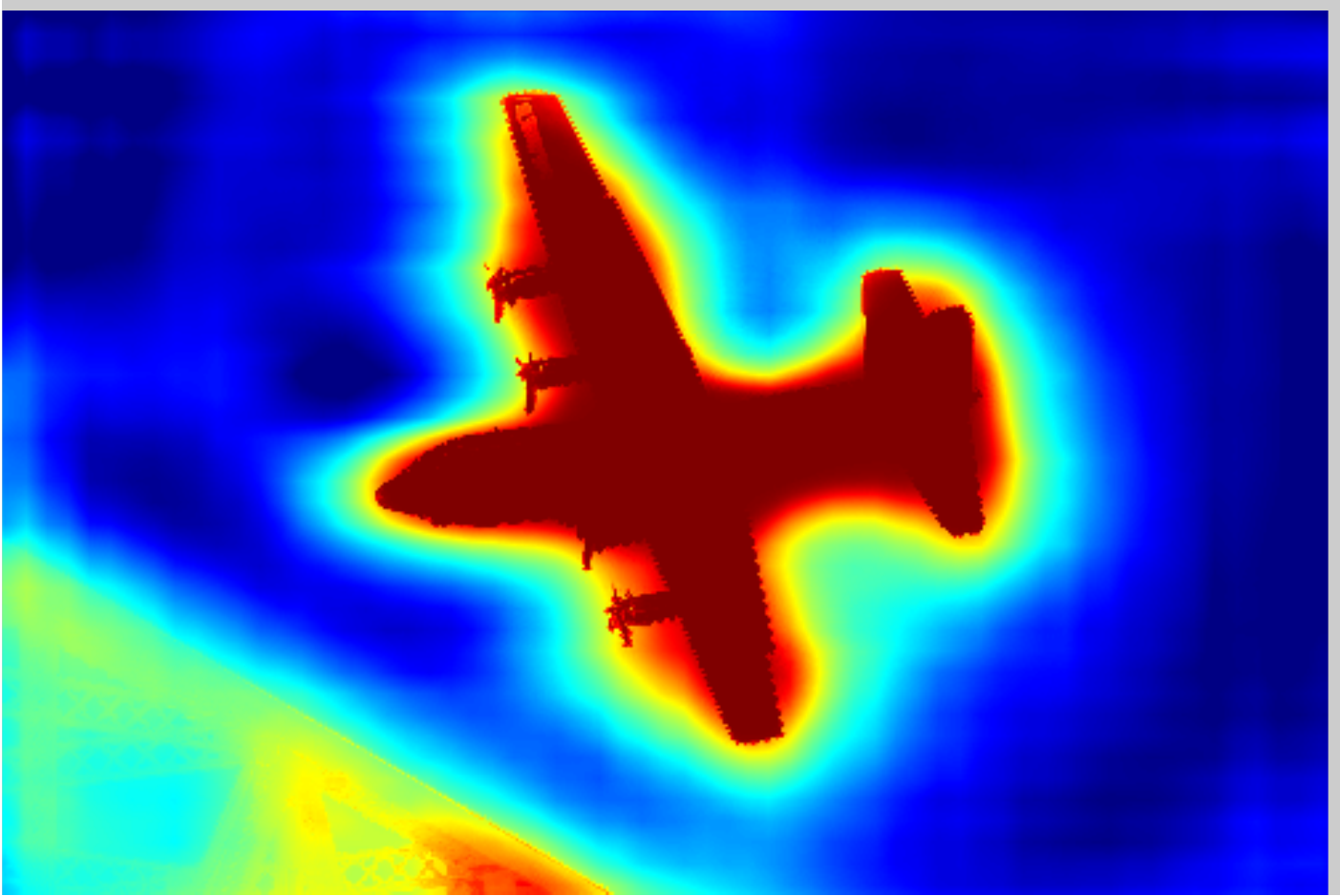} & 
    \includegraphics[width=0.16\linewidth]{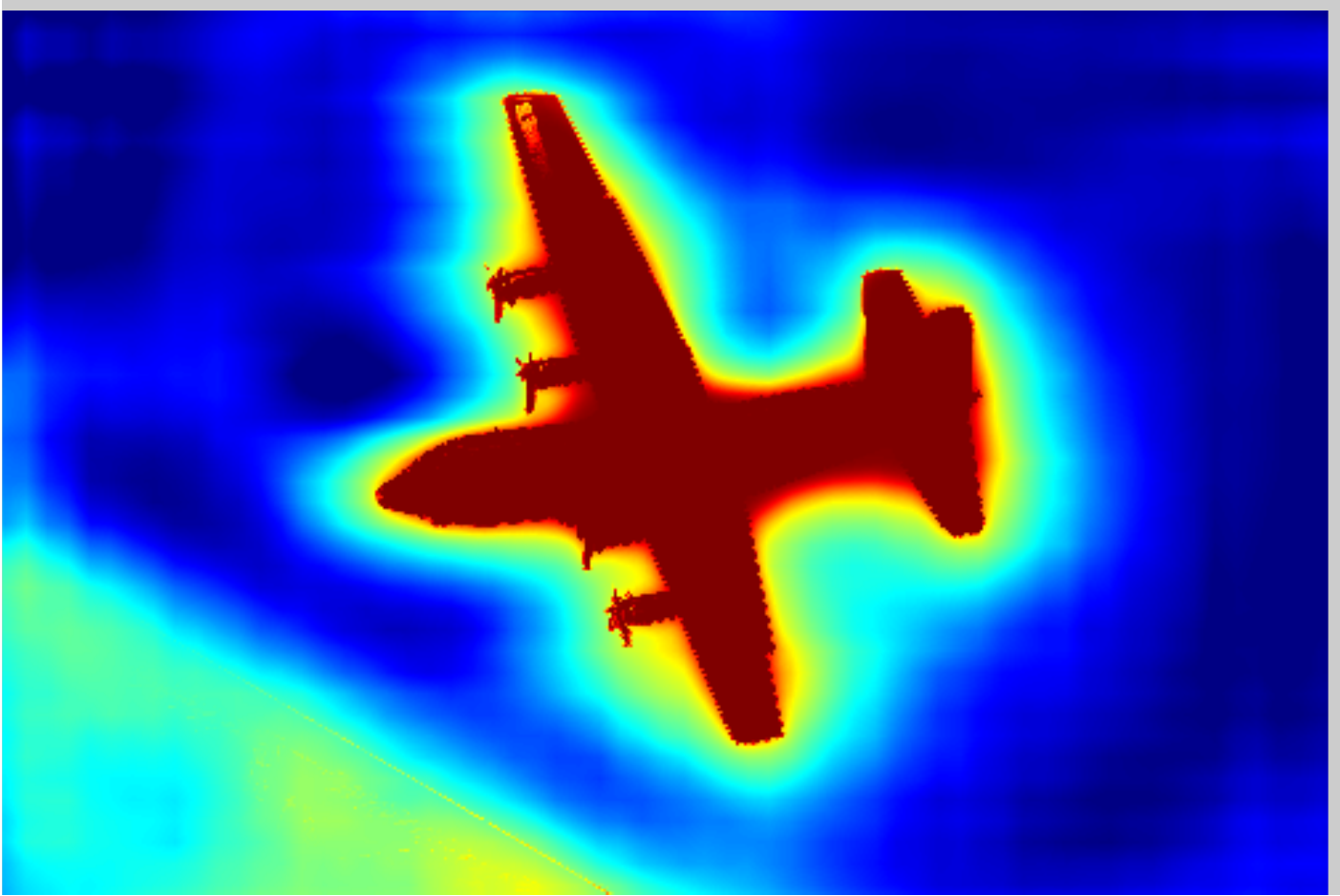} & 
    \includegraphics[width=0.16\linewidth]{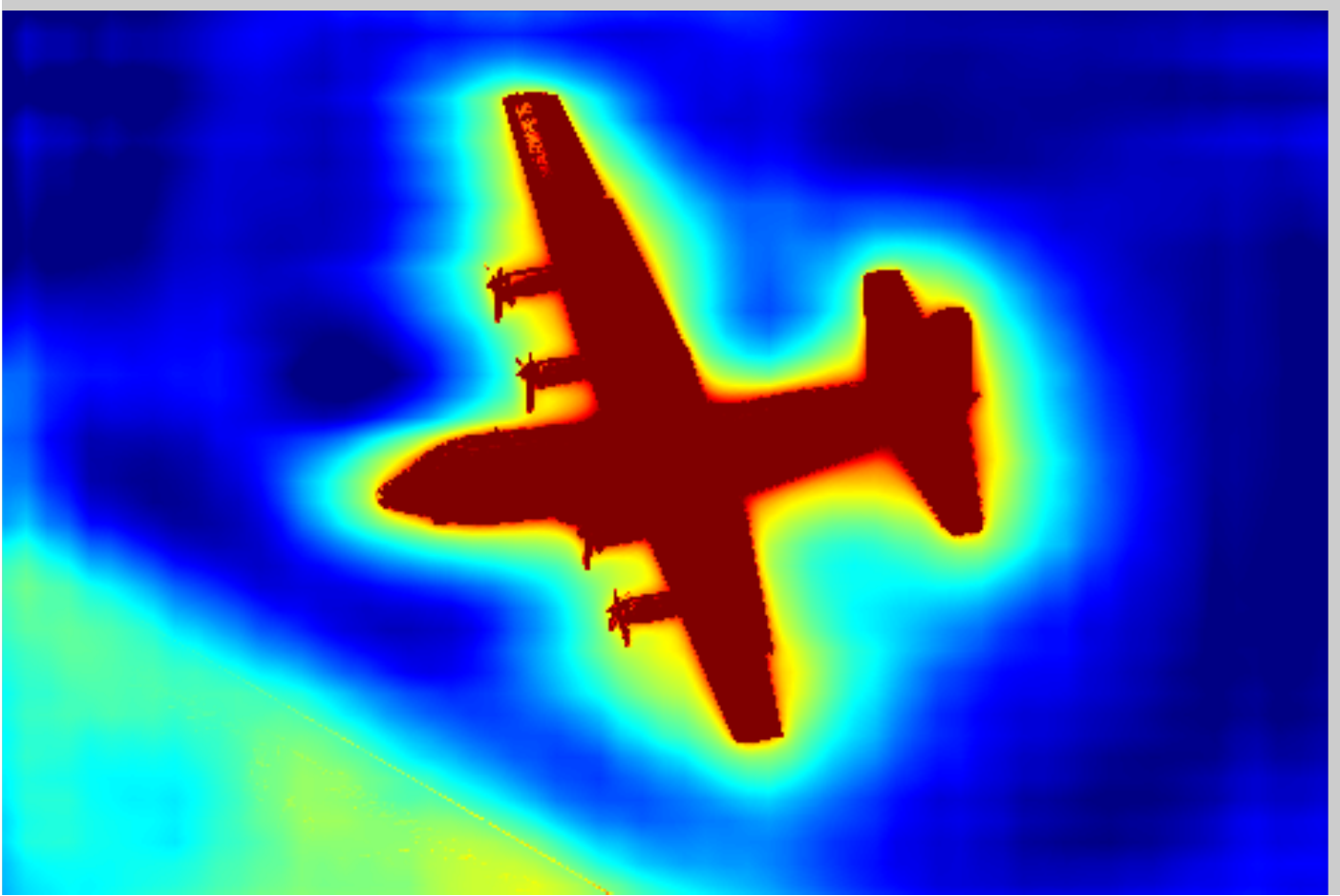} \\
    \includegraphics[width=0.16\linewidth]{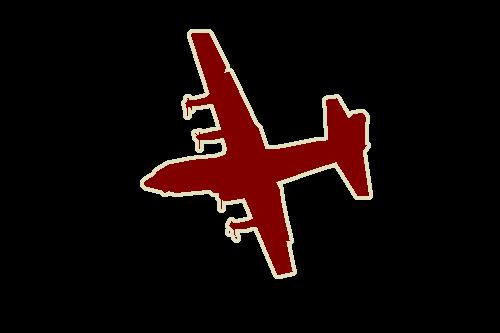} & 
    \includegraphics[width=0.16\linewidth]{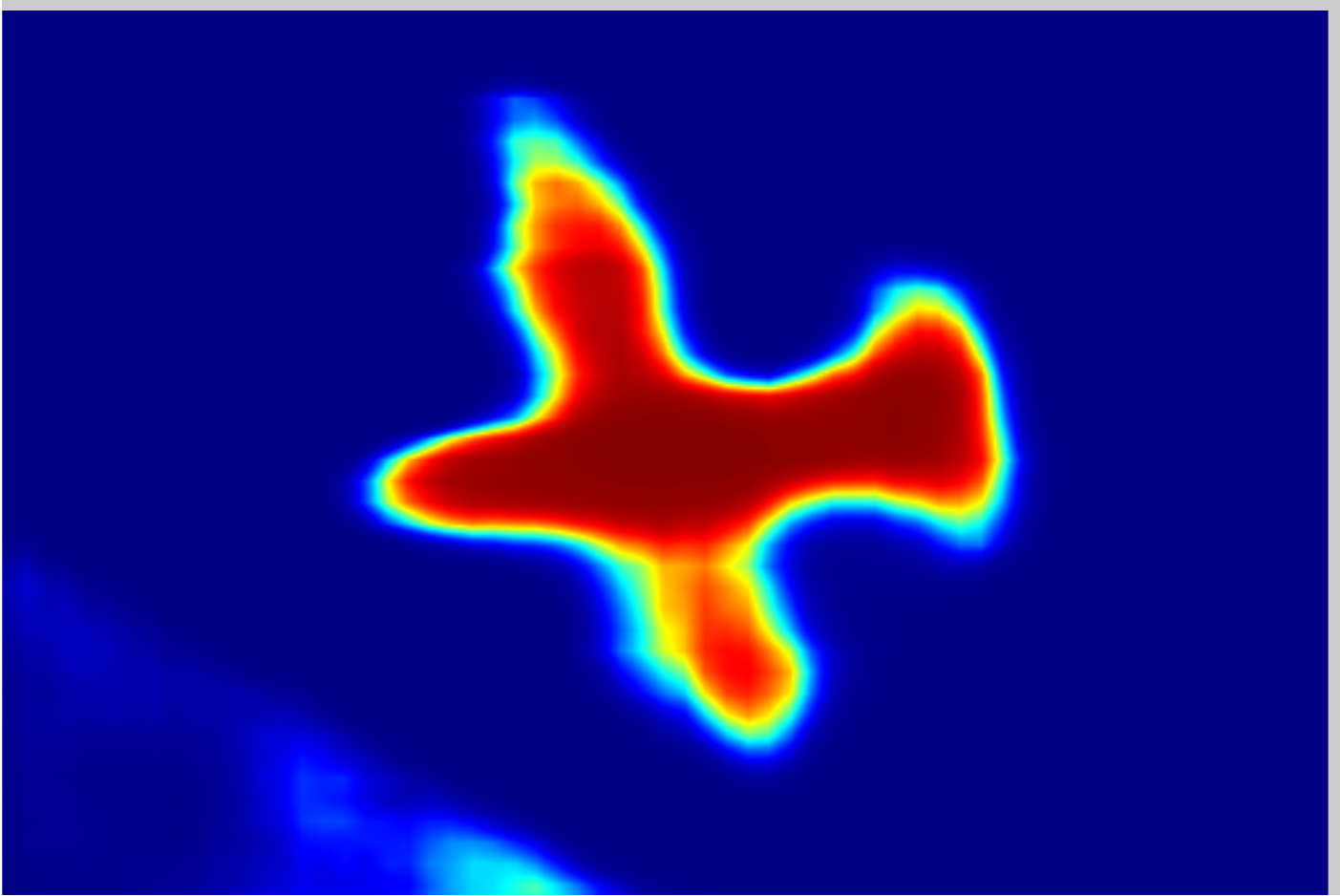} & 
    \includegraphics[width=0.16\linewidth]{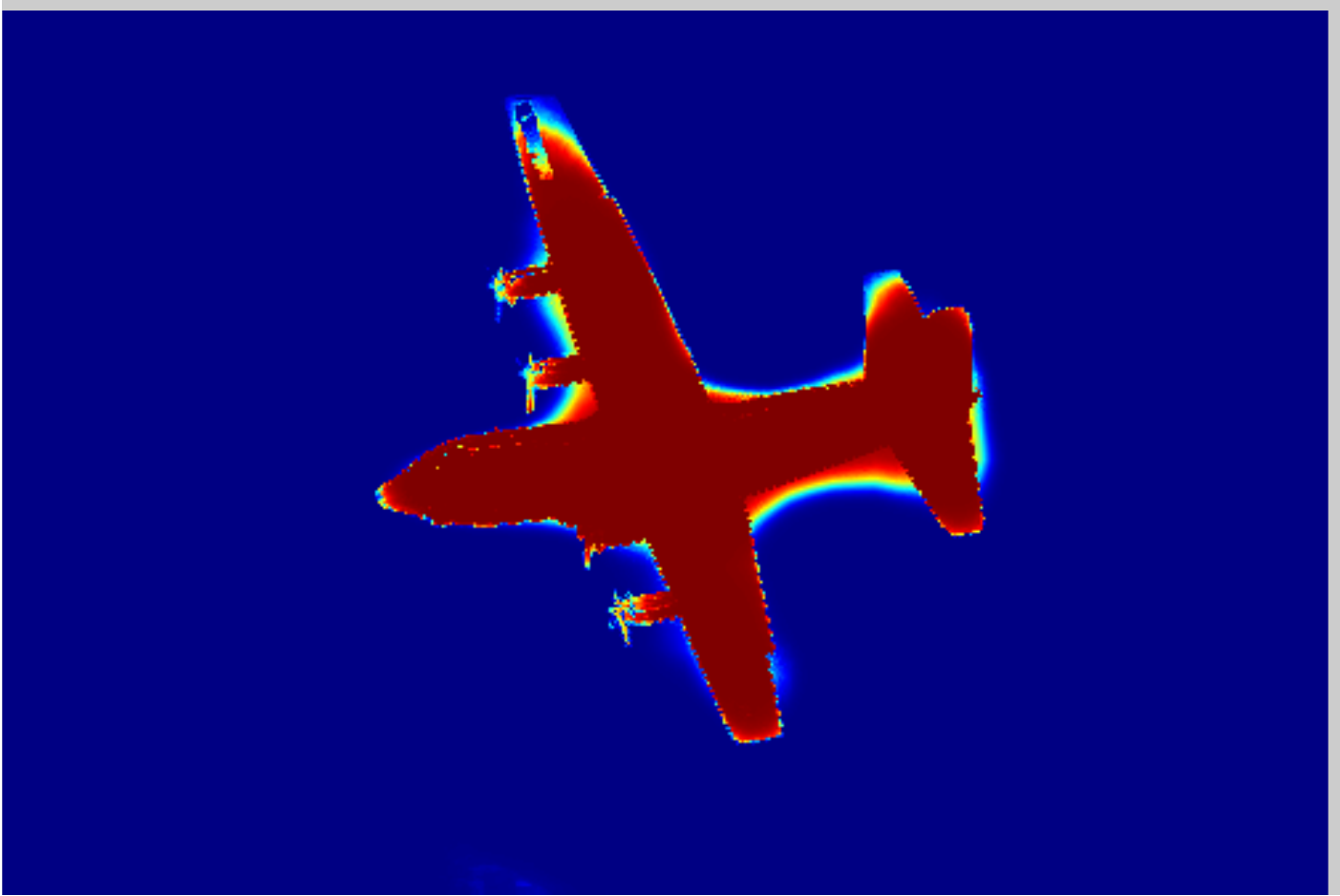} & 
    \includegraphics[width=0.16\linewidth]{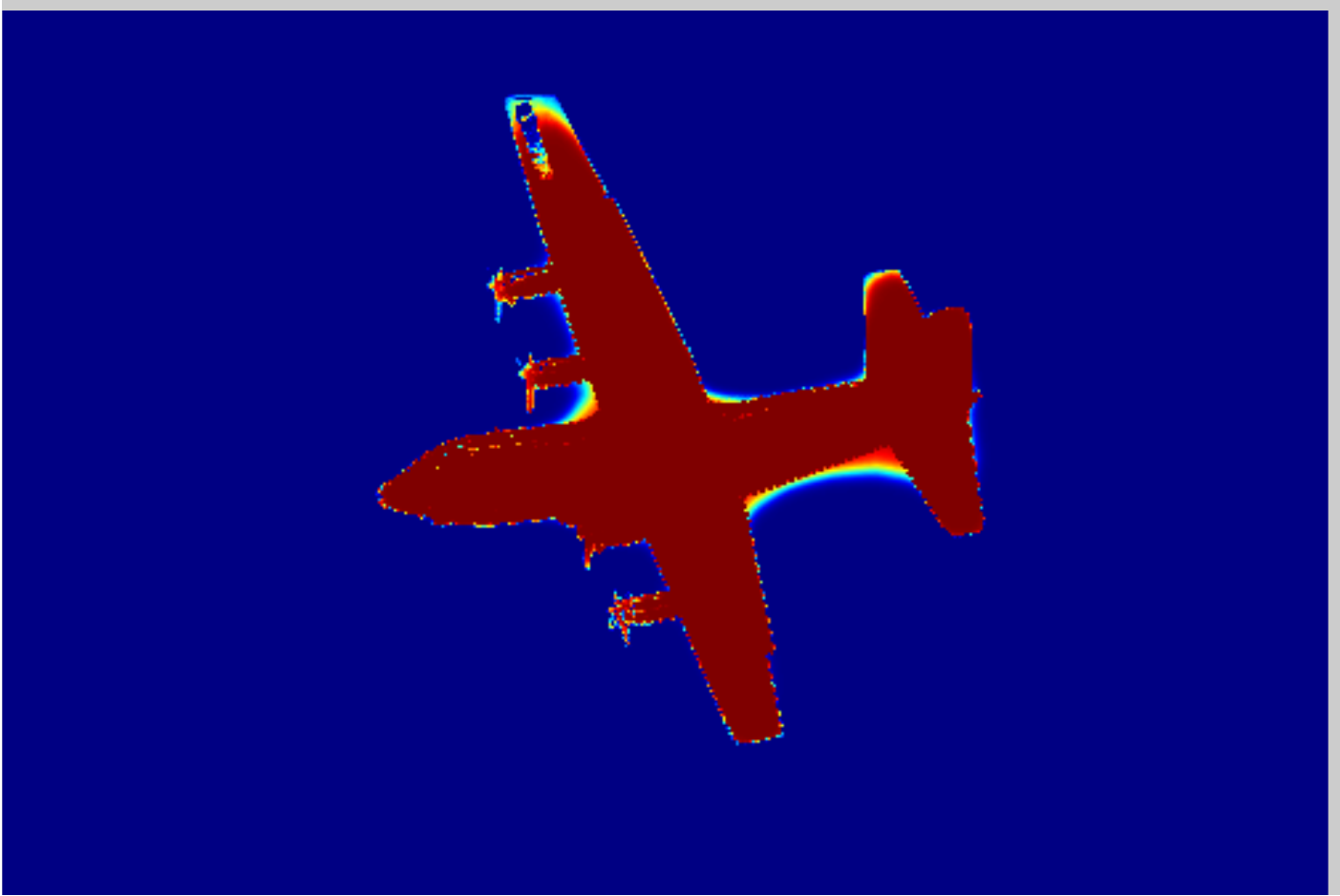} & 
    \includegraphics[width=0.16\linewidth]{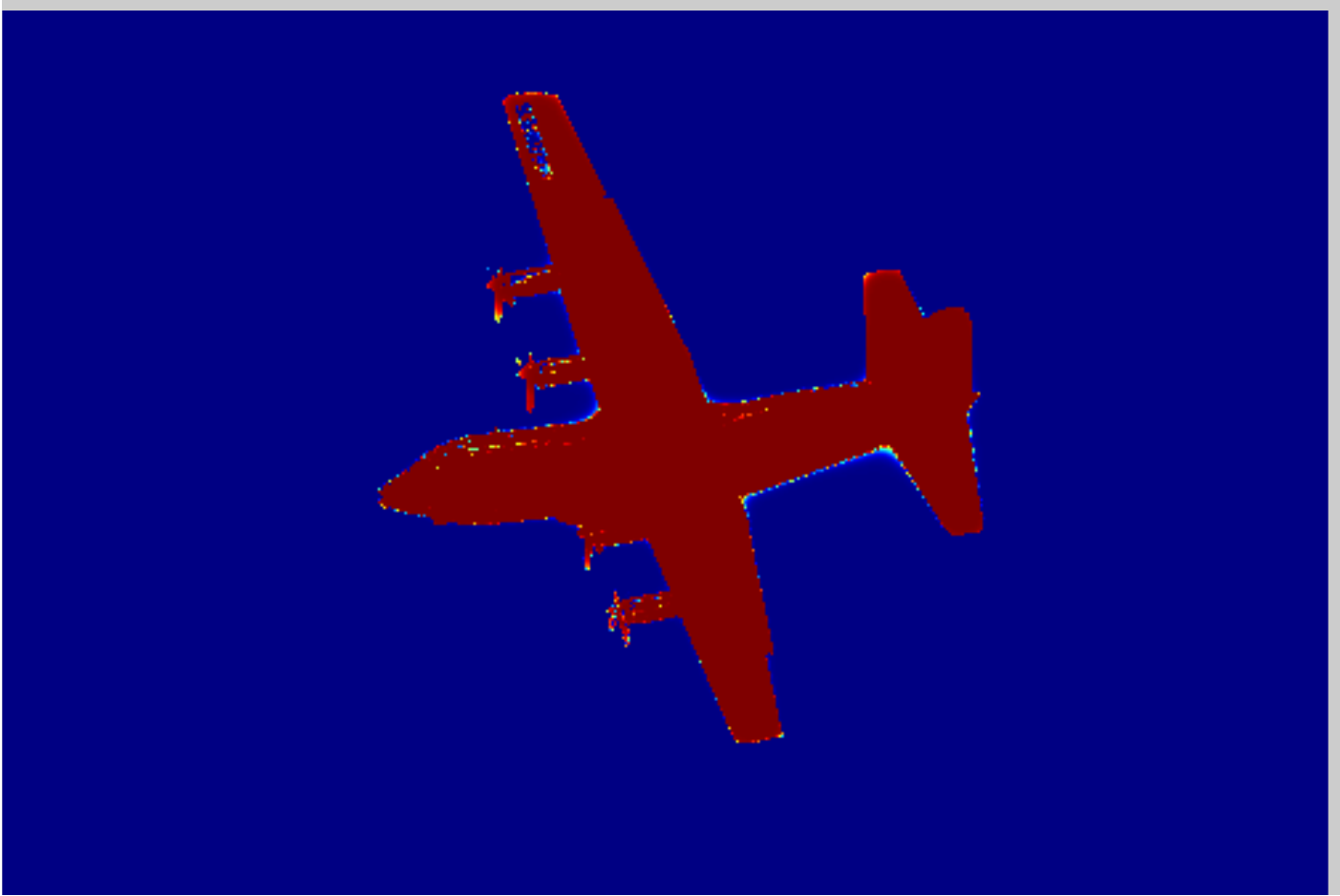} \\
    {\tiny Image/G.T.} & {\tiny DCNN output} & {\tiny CRF Iteration 1} & {\tiny CRF Iteration 2} & {\tiny CRF Iteration 10} \\
  \end{tabular}
  \caption{Score map (input before softmax function) and belief map (output of
    softmax function) for Aeroplane. We show the score (1st row) and belief (2nd row)
    maps after each mean field iteration. The output of last DCNN layer is used as
    input to the mean field inference.}
  \label{fig:score-maps}
\end{figure}

A trade-off between localization accuracy and classification performance seems to be inherent in DCNNs:
deeper models with multiple max-pooling layers have
proven most successful in classification tasks, however the increased
invariance and the large receptive fields of top-level nodes can only yield smooth responses.
As illustrated in \figref{fig:score-maps}, DCNN score maps can
 predict the presence and rough position of objects but
cannot really delineate their borders. 

Previous work has pursued two directions to address this localization challenge.
The first approach is to harness information from multiple layers in the
convolutional network in order to better estimate the object boundaries
\cite{hariharan2014hypercolumns, long2014fully, eigen2014predicting}. The
second  is to employ a super-pixel representation, essentially
delegating the localization task to a low-level segmentation method
\cite{mostajabi2014feedforward}.



We pursue an alternative direction based on coupling the recognition
capacity of DCNNs and the fine-grained localization accuracy of fully connected
CRFs and show that it is remarkably successful in addressing the localization
challenge, producing accurate semantic segmentation results and recovering
object boundaries at a level of detail that is well beyond the reach of existing
methods. This direction has been extended by several follow-up papers
\cite{papandreou2015weakly, schwing2015fully, zheng2015conditional,
  dai2015boxsup, noh2015learning, liu2015semantic, lin2015efficient,
  chen2015attention, chen2015semantic}, since the first version of our work
was published \cite{chen2014semantic}.

Traditionally, conditional random fields (CRFs) have been employed to smooth
noisy segmentation maps \cite{rother2004grabcut, kohli2009robust}. Typically
these models  couple neighboring nodes, favoring
same-label assignments to spatially proximal pixels. Qualitatively, the
primary function of these short-range CRFs is to clean up the spurious
predictions of weak classifiers built on top of local hand-engineered features.

Compared to these weaker classifiers, modern DCNN architectures such as
the one we use in this work produce score maps and semantic label
predictions which are qualitatively different. As illustrated in
\figref{fig:score-maps}, the score maps are typically quite smooth and
produce homogeneous classification results. In this regime, using short-range
CRFs can be detrimental, as our goal should be to recover detailed local
structure rather than further smooth it. Using contrast-sensitive potentials
\cite{rother2004grabcut} in conjunction to local-range CRFs can potentially
improve localization but still miss thin-structures and typically requires
solving an expensive discrete optimization problem.

To overcome these limitations of short-range CRFs, we integrate into our system
the fully connected CRF model of \cite{krahenbuhl2011efficient}. The model
employs the energy function
\begin{align}
  E(\boldsymbol{x}) = \sum_i \theta_i(x_i) + \sum_{ij} \theta_{ij}(x_i, x_j)
\end{align}
where $\boldsymbol{x}$ is the label assignment for pixels. We use as unary
potential $\theta_i(x_i) = - \log P(x_i)$, where $P(x_i)$ is the label
assignment probability at pixel $i$ as computed by a DCNN. The pairwise
potential has a form that allows for efficient inference while using a fully-connected graph, i.e.
when  connecting 
all pairs of image pixels, $i,j$. In particular, as in \cite{krahenbuhl2011efficient},  we use the following expression:
\begin{gather}
 \hspace{-.2cm}\theta_{ij}(x_i, x_j) \!=\! \mu(x_i,x_j)\!\left[w_1 \exp \Big(\!-\!\frac{||p_i-p_j||^2}{2\sigma_\alpha^2} \!-\!\frac{||I_i-I_j||^2}{2\sigma_\beta^2}\! \Big)\right. \nonumber\\
\left. + w_2 \exp \Big(-\frac{||p_i-p_j||^2}{2\sigma_\gamma^2}\Big)\right]\label{eq:fully_crf}
\end{gather}
where
$\mu(x_i,x_j)= 1 \text{ if } x_i \neq x_j$, and zero otherwise, which, as in the Potts model, means that only nodes with distinct labels are penalized.  The remaining expression uses two Gaussian kernels in different feature spaces; the first, `bilateral' kernel
depends on both pixel positions (denoted as $p$) and
RGB color (denoted as $I$), and the second kernel only depends on pixel
positions. The hyper parameters $\sigma_\alpha$, $\sigma_\beta$ and
$\sigma_\gamma$ control the scale of Gaussian kernels. The  first kernel forces pixels with similar color and position to have similar labels, while the second kernel only considers spatial proximity when enforcing smoothness. 


Crucially, this model is amenable to efficient approximate probabilistic
inference \cite{krahenbuhl2011efficient}. The message passing updates under a
fully decomposable mean field approximation $b(\boldsymbol{x}) = \prod_i
b_i(x_i)$ can be expressed as Gaussian convolutions in bilateral
space. High-dimensional filtering algorithms \cite{adams2010fast}
significantly speed-up this computation resulting in an algorithm that is very
fast in practice, requiring less that 0.5 sec on average for Pascal VOC images using the
publicly available implementation of \cite{krahenbuhl2011efficient}.

\section{Experimental Results}
\label{sec:experiments}

We finetune the model weights of the Imagenet-pretrained VGG-16 or ResNet-101
networks to adapt them to the semantic segmentation task in a straightforward
fashion, following the procedure of \cite{long2014fully}. We replace the
1000-way Imagenet classifier in the last layer with a classifier having as many
targets as the number of semantic classes of our task (including the background,
if applicable). Our loss function is the sum of cross-entropy terms for each
spatial position in the CNN output map (subsampled by 8 compared to the original
image). All positions and labels are equally weighted in the overall loss
function (except for unlabeled pixels which are ignored). Our targets are the
ground truth labels (subsampled by 8). We optimize the objective function with
respect to the weights at all network layers by the standard SGD procedure of
\cite{KrizhevskyNIPS2013}. We decouple the DCNN and CRF training stages,
assuming the DCNN unary terms are fixed when setting the CRF parameters.

We evaluate the proposed models on four challenging datasets: PASCAL VOC 2012,
PASCAL-Context, PASCAL-Person-Part, and Cityscapes. We first report the main
results of our conference version \cite{chen2014semantic} on PASCAL VOC 2012,
and move forward to latest results on all datasets.


\subsection{PASCAL VOC 2012}

\textbf{Dataset:} The PASCAL VOC 2012 segmentation benchmark
\cite{everingham2014pascal} involves 20 foreground object classes and one
background class. The original dataset contains $1,464$ (\textit{train}),
$1,449$ (\textit{val}), and $1,456$ (\textit{test}) pixel-level labeled
images for training, validation, and testing, respectively. The dataset is
augmented by the extra annotations provided by \cite{hariharan2011semantic},
resulting in $10,582$ (\textit{trainaug}) training images. The performance
is measured in terms of pixel intersection-over-union (IOU) averaged across
the 21 classes.

\subsubsection{Results from our conference version}

We employ the VGG-16 network pre-trained on Imagenet, adapted for semantic
segmentation as described in Section~\ref{sec:convnet-hole}. We use a
mini-batch of 20 images and initial learning rate of $0.001$ ($0.01$
for the final classifier layer), multiplying the learning rate by 0.1 every
2000 iterations. We use momentum of $0.9$ and weight decay of $0.0005$.

After the DCNN has been fine-tuned on \textit{trainaug}, we cross-validate the
CRF parameters along the lines of \cite{krahenbuhl2011efficient}. We use default
values of $w_2 = 3$ and $\sigma_\gamma = 3$ and we search for the best values of
$w_1$, $\sigma_\alpha$, and $\sigma_\beta$ by cross-validation on 100 images
from \textit{val}. We employ a coarse-to-fine search scheme. The initial search
range of the parameters are $w_1 \in [3:6]$, $\sigma_\alpha \in [30:10:100]$ and
$\sigma_\beta \in [3:6]$ (MATLAB notation), and then we refine the search step
sizes around the first round's best values. We employ 10 mean field iterations.

\begin{table}[!t]
  \centering
  \addtolength{\tabcolsep}{2.5pt}
  \scalebox{0.97}{
  \begin{tabular}{c c c | c c | c}
    \toprule[0.2em]
    {\bf Kernel} & {\bf Rate} & {\bf FOV} & {\bf Params} & {\bf Speed} & {\bf bef/aft CRF} \\
    \toprule[0.2em]
    \by{7}{7}         &    4  & 224 & 134.3M & 1.44 & 64.38 / 67.64 \\
    \by{4}{4}         &    4  & 128 & 65.1M  & 2.90 & 59.80 / 63.74 \\
    \by{4}{4}         &    8  & 224 & 65.1M  & 2.90 & 63.41 / 67.14 \\
    \by{3}{3}         &   12  & 224 & 20.5M  & 4.84 & 62.25 / 67.64 \\
    \bottomrule[0.1em]
  \end{tabular}
  }
  \caption{Effect of Field-Of-View by adjusting the kernel size and atrous
    sampling rate $r$ at `fc6' layer. We show number of model parameters,
    training speed (img/sec), and \textit{val} set mean IOU before and after
    CRF. DeepLab-LargeFOV (kernel size \by{3}{3}, $r = 12$) strikes the best
    balance.}
  \label{tab:fov}
\end{table}

\textbf{Field of View and CRF:}
 In \tabref{tab:fov}, we report experiments with DeepLab model variants that use
different field-of-view sizes, obtained by adjusting the kernel size and atrous
sampling rate $r$ in the `fc6' layer, as described in \secref{sec:convnet-hole}.
We start with a direct adaptation of VGG-16 net, using the original \by{7}{7}
kernel size and $r = 4$ (since we use no stride for the last two max-pooling
layers). This model yields performance of $67.64\%$ after CRF, but is relatively
slow ($1.44$ images per second during training). We have improved model speed to
$2.9$ images per second by reducing the kernel size to \by{4}{4}. We have
experimented with two such network variants with smaller ($r = 4$) and larger
($r = 8$) FOV sizes; the latter one performs better. Finally, we employ kernel
size \by{3}{3} and even larger atrous sampling rate ($r = 12$), also making the
network thinner by retaining a random subset of 1,024 out of the 4,096 filters
in layers `fc6' and `fc7'. The resulting model, DeepLab-CRF-LargeFOV, matches
the performance of the direct VGG-16 adaptation (\by{7}{7} kernel size, $r = 4$).
At the same time, DeepLab-LargeFOV is $3.36$ times faster and has significantly
fewer parameters (20.5M instead of 134.3M).

The CRF substantially boosts performance of all model variants, offering a 3-5\%
absolute increase in mean IOU.

\textbf{Test set evaluation:} We have evaluated our DeepLab-CRF-LargeFOV model
on the PASCAL VOC 2012 official \textit{test} set. It achieves $70.3\%$ mean IOU
performance.

\begin{figure*}[!htbp]
  \centering
  \scalebox{0.55} {
  \begin{tabular}{c}
    \includegraphics[width=1.8\linewidth]{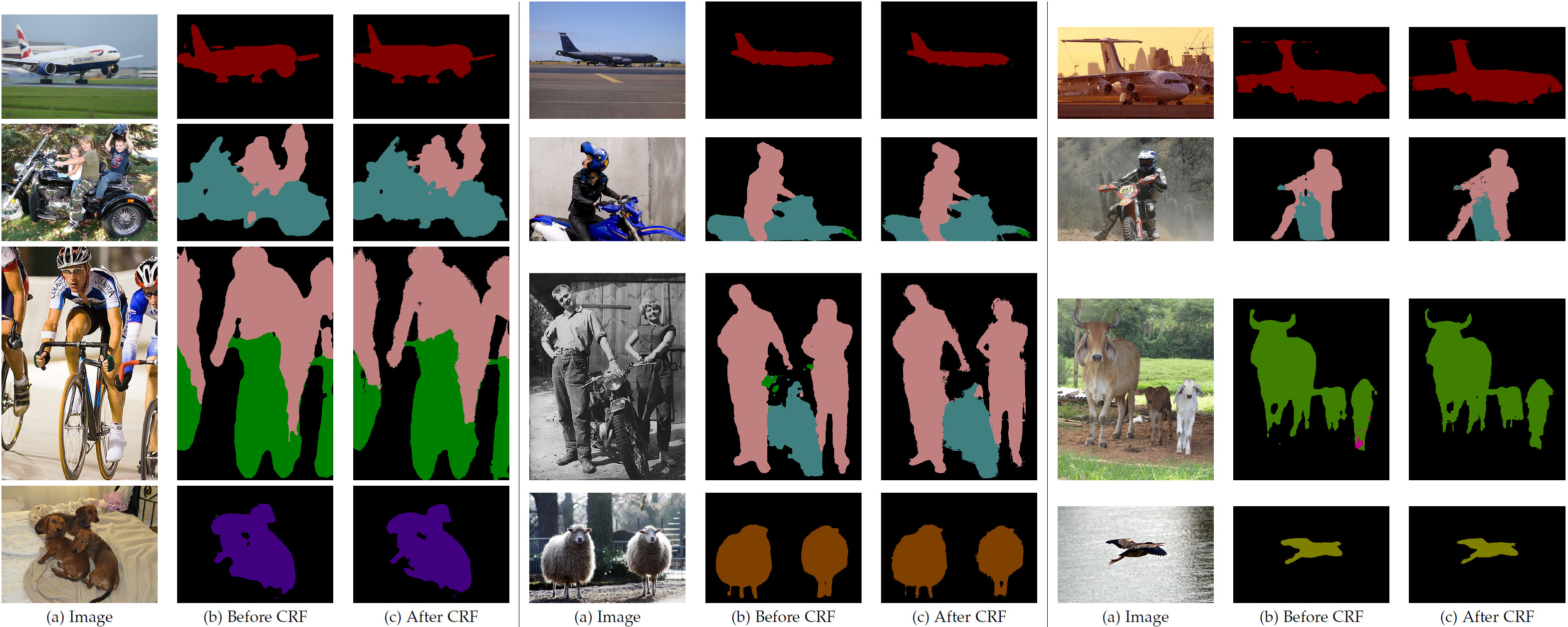}\\
  \end{tabular}
  }
  \caption{PASCAL VOC 2012 \textit{val} results. Input image
    and our DeepLab results before/after CRF.}
  \label{fig:ValResults}
\end{figure*}

\subsubsection{Improvements after conference version of this work}
After the conference version of this work \cite{chen2014semantic}, we have
pursued three main improvements of our model, which we discuss below:
(1) different learning policy during training, (2) atrous spatial pyramid
pooling, and (3) employment of deeper networks and multi-scale processing.

\textbf{Learning rate policy:} We have explored different learning rate
policies when training DeepLab-LargeFOV. Similar to \cite{liu2015parsenet},
we also found that employing a ``poly'' learning rate policy (the learning
rate is multiplied by $(1-\frac{iter}{max\_iter})^{power}$) is more effective
than ``step'' learning rate (reduce the learning rate at a fixed step size).
As shown in \tabref{tab:val_poly}, employing ``poly'' (with $power = 0.9$)
and using the same batch size and same training iterations yields 1.17\% better
performance than employing ``step'' policy. Fixing the batch size and increasing
the training iteration to 10K improves the performance to 64.90\% (1.48\% gain);
however, the total training time increases due to more training iterations. We
then reduce the batch size to 10 and found that comparable performance is still
maintained (64.90\% \vs 64.71\%). In the end, we employ batch size = 10 and
20K iterations in order to maintain similar training time as previous ``step''
policy. Surprisingly, this gives us the performance of 65.88\% (3.63\%
improvement over ``step'') on \textit{val}, and 67.7\% on \textit{test},
compared to 65.1\% of the original ``step'' setting for DeepLab-LargeFOV before
CRF. We employ the ``poly'' learning rate policy for all experiments reported in
the rest of the paper.

\begin{table}[!t]
  \centering
  \addtolength{\tabcolsep}{2.5pt}
  \begin{tabular}{c c c c}
    \toprule[0.2 em]
    {\bf Learning policy} & {\bf Batch size} & {\bf Iteration} & {\bf mean IOU} \\
    \toprule[0.2em]
    step & 30 & 6K & 62.25 \\
    \midrule
    poly & 30 & 6K & 63.42 \\
    poly & 30 & 10K & 64.90 \\
    poly & 10 & 10K & 64.71 \\
    poly & 10 & 20K & 65.88 \\
    \bottomrule[0.1em]
  \end{tabular}
  \caption{PASCAL VOC 2012 \textit{val} set results (\%) (before CRF) as
    different learning hyper parameters vary. Employing ``poly'' learning
    policy is more effective than ``step'' when training DeepLab-LargeFOV.}
  \label{tab:val_poly}
\end{table}

\begin{figure}[!t]
  \centering
  \scalebox{0.85}{
  \begin{tabular}{c c}
    \includegraphics[width=0.2\linewidth]{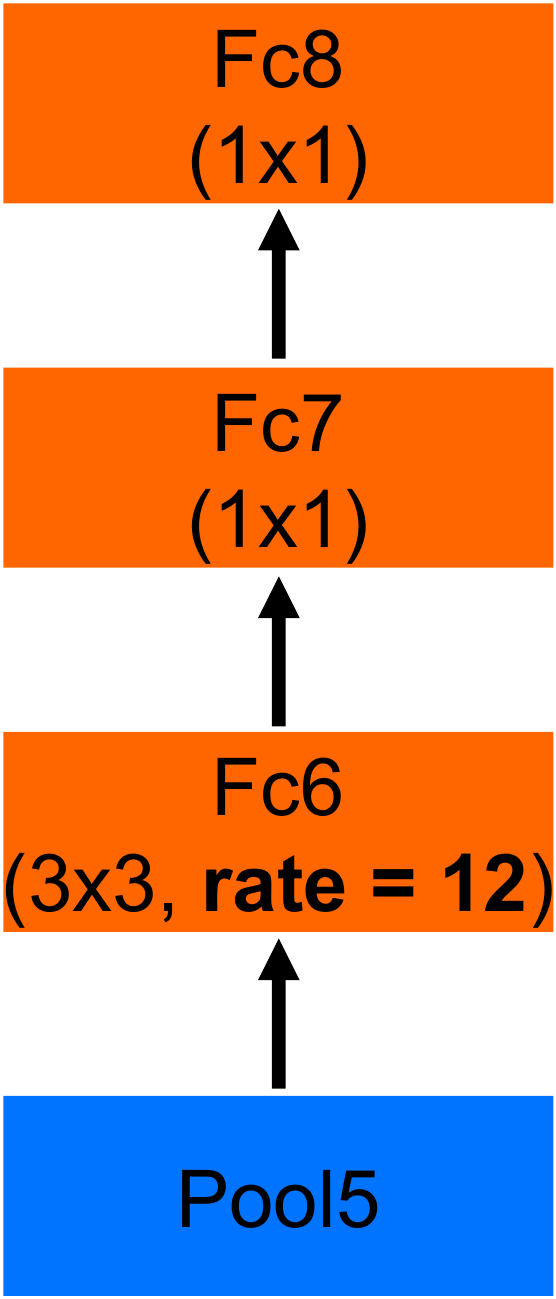} &
    \includegraphics[height=4.5cm]{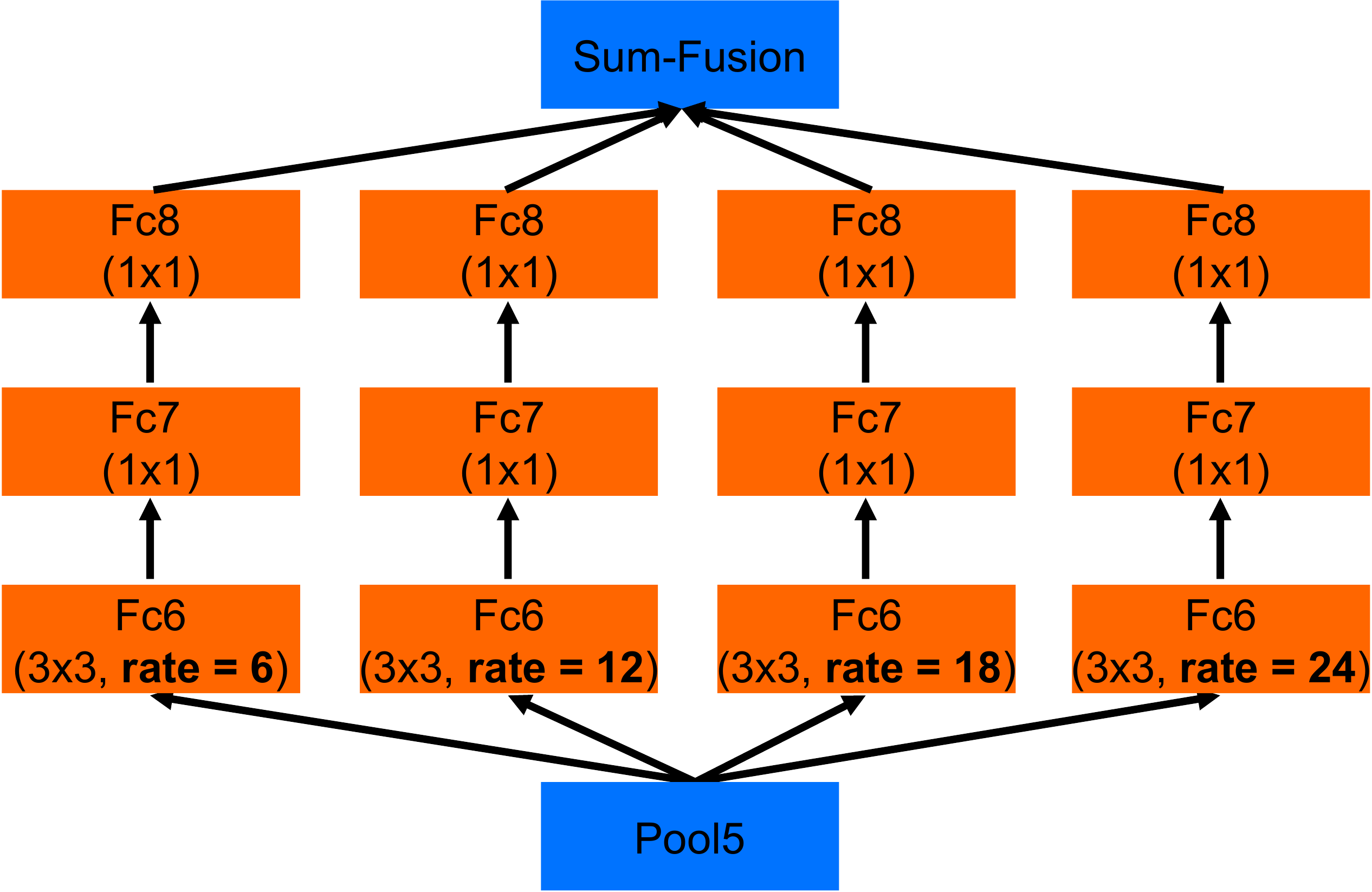} \\
    {\scriptsize (a) DeepLab-LargeFOV} &
    {\scriptsize (b) DeepLab-ASPP} \\
  \end{tabular}
  }
  \caption{DeepLab-ASPP employs multiple filters with different rates to capture objects and context at multiple
    scales.}
  \label{fig:diff_hole}
\end{figure}

\begin{figure}
  \centering
  \begin{tabular}{c c c c}
    \includegraphics[width=0.21\linewidth]{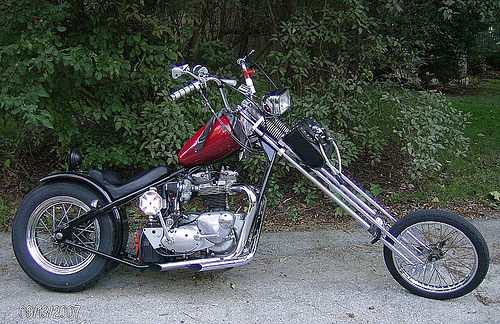} &
    \includegraphics[width=0.21\linewidth]{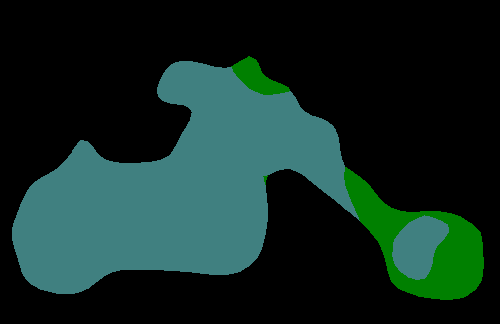} &
    \includegraphics[width=0.21\linewidth]{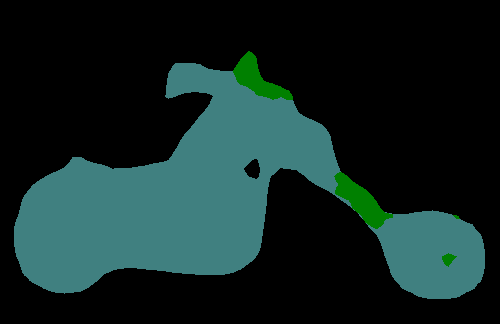} &    
    \includegraphics[width=0.21\linewidth]{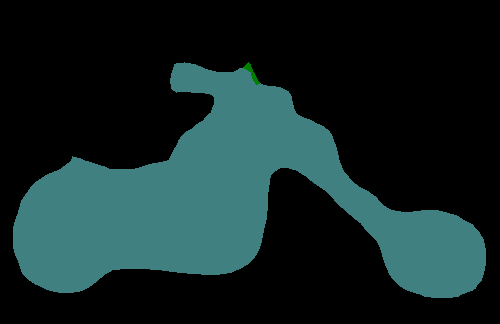} \\
    \includegraphics[width=0.21\linewidth]{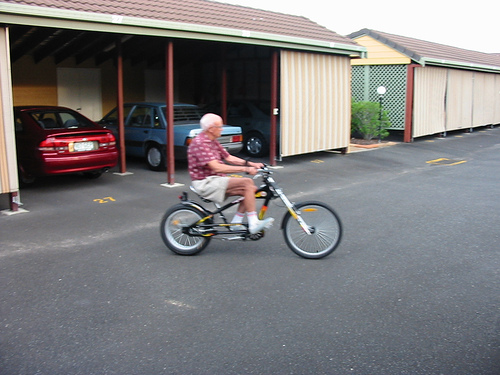} &
    \includegraphics[width=0.21\linewidth]{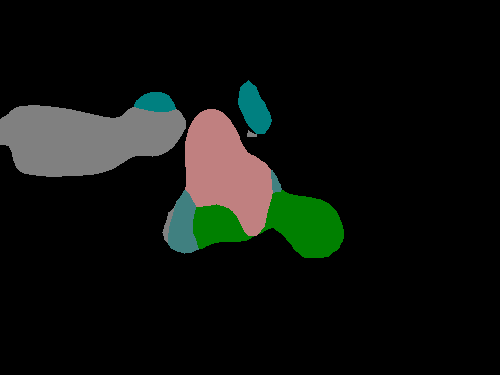} &
    \includegraphics[width=0.21\linewidth]{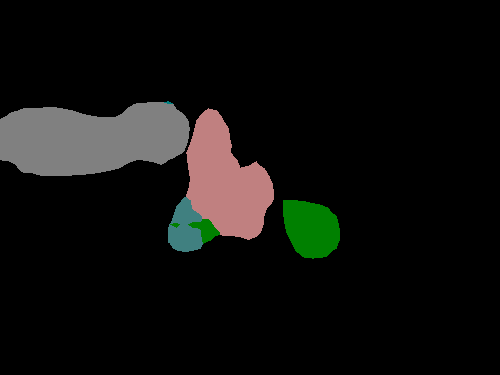} &    
    \includegraphics[width=0.21\linewidth]{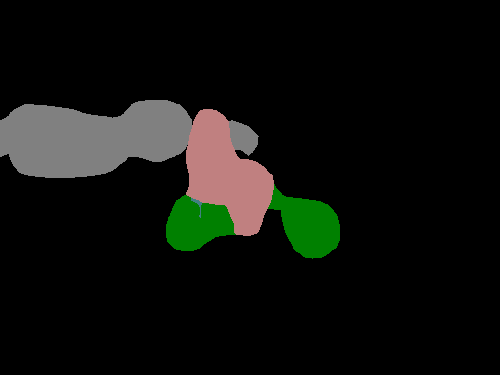} \\
    (a) Image &
    (b) LargeFOV &
    (c) ASPP-S &
    (d) ASPP-L \\
  \end{tabular}
  \caption{Qualitative segmentation results with ASPP compared to the baseline
    LargeFOV model. The \textbf{ASPP-L} model, employing multiple {\it large}
    FOVs can successfully capture objects as well as image context at multiple
    scales.}
  \label{fig:aspp}
\end{figure}

\textbf{Atrous Spatial Pyramid Pooling:} We have experimented with the proposed
Atrous Spatial Pyramid Pooling (ASPP) scheme, described in
\secref{sec:convnet-hole}. As shown in \figref{fig:diff_hole}, ASPP for VGG-16
employs several parallel fc6-fc7-fc8 branches. They all use \by{3}{3} kernels
but different atrous rates $r$ in the `fc6' in order to capture objects of
different size. In \tabref{tab:vgg_mfov}, we report results with several
settings:
(1) Our baseline LargeFOV model, having a single branch with $r=12$,
(2) ASPP-S, with four branches and smaller atrous rates ($r$ = \{2, 4, 8, 12\}), and
(3) ASPP-L, with four branches and larger rates ($r$ = \{6, 12, 18, 24\}).
For each variant we report results before and after CRF.
As shown in the table, ASPP-S yields 1.22\% improvement over the baseline
LargeFOV before CRF. However, after CRF both LargeFOV and ASPP-S perform similarly.
On the other hand, ASPP-L yields consistent improvements over the baseline LargeFOV
both before and after CRF. We evaluate on \textit{test} the proposed ASPP-L + CRF
model, attaining 72.6\%. We visualize the effect of the different schemes in
\figref{fig:aspp}.

\begin{table}[!t]
  \centering
  \addtolength{\tabcolsep}{0pt}
  \begin{tabular} {c | c c }
    \toprule[0.2em]
    {\bf Method} & {\bf before CRF} & {\bf after CRF} \\
    \toprule[0.2em]
    LargeFOV & 65.76 & 69.84 \\
    ASPP-S   & 66.98 & 69.73 \\
    ASPP-L   & 68.96 & 71.57 \\
    \bottomrule[0.1em]
  \end{tabular}
  \caption{Effect of ASPP on PASCAL VOC 2012 \textit{val} set
    performance (mean IOU) for VGG-16 based DeepLab model.
    {\bf LargeFOV}: single branch, $r = 12$.
    {\bf ASPP-S}: four branches, $r$ = \{2, 4, 8, 12\}.
    {\bf ASPP-L}: four branches, $r$ = \{6, 12, 18, 24\}.}
  \label{tab:vgg_mfov}
\end{table}

\begin{table}[!t]
  \centering
  \addtolength{\tabcolsep}{-1pt}
  \begin{tabular} {c c c c c c | c}
    \toprule[0.2em]
    {\bf MSC} & {\bf COCO} & {\bf Aug} & {\bf LargeFOV} & {\bf ASPP} & {\bf CRF} & {\bf mIOU} \\
    \toprule[0.2em]
    & & & & & & 68.72 \\
    \checkmark & & & & & & 71.27 \\
    \checkmark & \checkmark & & & & & 73.28 \\
    \checkmark & \checkmark & \checkmark & & & & 74.87 \\
    \checkmark & \checkmark & \checkmark & \checkmark & & & 75.54 \\
    \checkmark & \checkmark & \checkmark & & \checkmark & & 76.35 \\
    \checkmark & \checkmark & \checkmark & & \checkmark & \checkmark & 77.69 \\
    \bottomrule[0.1em]
  \end{tabular}
  \caption{Employing ResNet-101 for DeepLab on PASCAL VOC 2012 {\it val} set.
    {\bf MSC}: Employing mutli-scale inputs with max fusion.
    {\bf COCO}: Models pretrained on MS-COCO.
    {\bf Aug}: Data augmentation by randomly rescaling inputs.}
  \label{tab:resnet_val}
\end{table}

\textbf{Deeper Networks and Multiscale Processing:} We have experimented
building DeepLab around the recently proposed residual net ResNet-101
\cite{he2015deep} instead of VGG-16. Similar to what we did for VGG-16 net,
we re-purpose ResNet-101 by atrous convolution, as described in
\secref{sec:convnet-hole}. On top of that, we adopt several other features,
following recent work of \cite{farabet2013learning, papandreou2015weakly,
  zheng2015conditional, liu2015semantic, lin2015efficient, chen2015attention,
  kokkinos2016pushing}: (1) Multi-scale inputs: We separately feed to the
DCNN images at scale = \{0.5, 0.75, 1\}, fusing their score maps by taking
the maximum response across scales for each position separately
\cite{chen2015attention}. (2) Models pretrained on MS-COCO
\cite{lin2014microsoft}. (3) Data augmentation by randomly scaling the input
images (from 0.5 to 1.5) during training. In \tabref{tab:resnet_val}, we
evaluate how each of these factors, along with LargeFOV and atrous spatial
pyramid pooling (ASPP), affects \textsl{val} set performance.
Adopting ResNet-101 instead of VGG-16 significantly improves DeepLab performance
(\eg, our simplest ResNet-101 based model attains 68.72\%, compared to 65.76\%
of our DeepLab-LargeFOV VGG-16 based variant, both before CRF). Multiscale
fusion \cite{chen2015attention} brings extra 2.55\% improvement, while
pretraining the model on MS-COCO gives another 2.01\% gain. Data augmentation
during training is effective (about 1.6\% improvement). Employing LargeFOV
(adding an atrous convolutional layer on top of ResNet, with \by{3}{3} kernel
and rate = 12) is beneficial (about 0.6\% improvement). Further 0.8\%
improvement is achieved by atrous spatial pyramid pooling (ASPP).
Post-processing our best model by dense CRF yields performance of 77.69\%.

\textbf{Qualitative results:} We provide qualitative visual comparisons of DeepLab's
results (our best model variant) before and after CRF in \figref{fig:ValResults}.
The visualization results obtained by DeepLab before CRF already yields excellent
segmentation results, while employing the CRF further improves the performance by
removing false positives and refining object boundaries.


\textbf{Test set results:} We have submitted the result of our final best model
to the official server, obtaining \textit{test} set performance of 79.7\%, as
shown in \tabref{tab:res_testset}. The model substantially outperforms previous
DeepLab variants (\eg, DeepLab-LargeFOV with VGG-16 net) and is currently the
top performing method on the PASCAL VOC 2012 segmentation leaderboard. 


\begin{table}[!th]
  \centering
  \addtolength{\tabcolsep}{2.5pt}
  \begin{tabular}{l | c}
    \toprule[0.2 em]
    {\bf Method} & {\bf mIOU} \\
    \toprule[0.2 em]
    DeepLab-CRF-LargeFOV-COCO \cite{papandreou2015weakly} & 72.7\\
    MERL\_DEEP\_GCRF \cite{Vemulapalli2016Gaussian} & 73.2 \\
    CRF-RNN \cite{zheng2015conditional} & 74.7 \\
    POSTECH\_DeconvNet\_CRF\_VOC \cite{noh2015learning} & 74.8 \\
    BoxSup \cite{dai2015boxsup} & 75.2 \\
    Context + CRF-RNN \cite{yu2015multi} & 75.3 \\
    $QO_4^{mres}$ \cite{chandra2016fast} & 75.5 \\
    DeepLab-CRF-Attention \cite{chen2015attention} & 75.7 \\
    CentraleSuperBoundaries++ \cite{kokkinos2016pushing} & 76.0 \\
    DeepLab-CRF-Attention-DT  \cite{chen2015semantic} & 76.3 \\
    H-ReNet + DenseCRF \cite{yan2016combining} & 76.8 \\
    LRR\_4x\_COCO \cite{ghiasi2016laplacian} & 76.8 \\
    DPN \cite{liu2015semantic} & 77.5 \\
    Adelaide\_Context \cite{lin2015efficient} & 77.8 \\
    Oxford\_TVG\_HO\_CRF \cite{arnab2015higher} & 77.9 \\
    Context CRF + Guidance CRF \cite{Shen2016Fast} & 78.1 \\
    Adelaide\_VeryDeep\_FCN\_VOC \cite{wu2016bridging} & 79.1 \\
    \midrule
    \href{http://host.robots.ox.ac.uk:8080/anonymous/FLHY8R.html}{DeepLab-CRF (ResNet-101)} & 79.7 \\
    \bottomrule[0.1 em]
  \end{tabular}
  \caption{Performance on PASCAL VOC 2012 {\it test} set. We have added some
    results from recent arXiv papers on top of the official leadearboard results.}
  \label{tab:res_testset}
\end{table}

\begin{figure}[!t]
  \centering
  \scalebox{0.85} {
  \begin{tabular}{c @{\hskip 5pt} c @{\hskip 5pt} c @{\hskip 5pt} c @{\hskip 5pt} c}


    \includegraphics[width=0.21\linewidth]{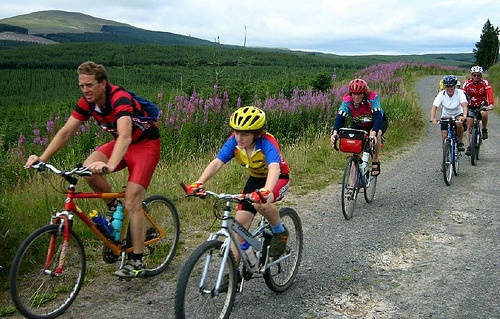} &
    \includegraphics[width=0.21\linewidth]{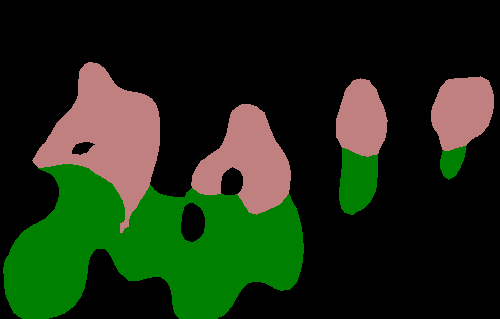} &
    \includegraphics[width=0.21\linewidth]{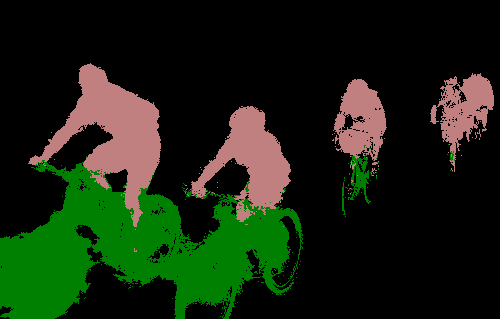} &
    \includegraphics[width=0.21\linewidth]{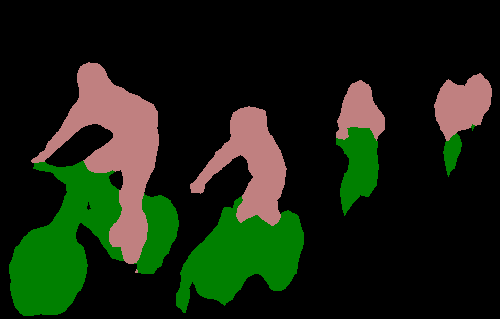} &
    \includegraphics[width=0.21\linewidth]{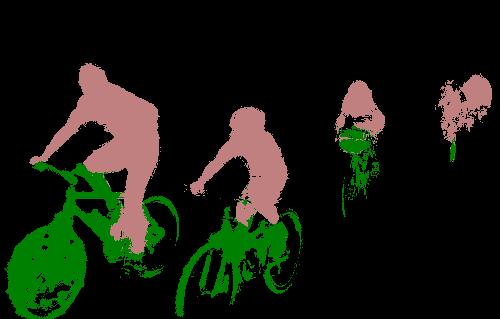} \\

    \includegraphics[width=0.21\linewidth]{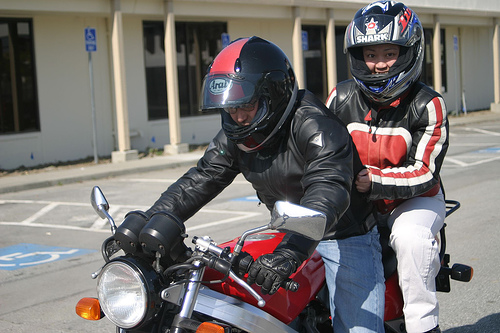} &
    \includegraphics[width=0.21\linewidth]{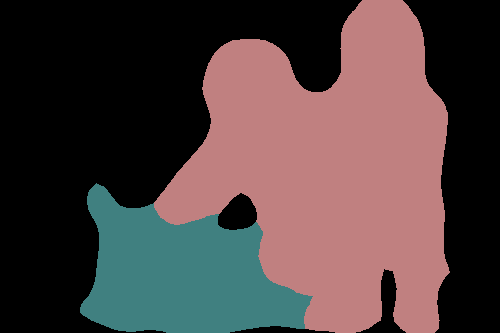} &
    \includegraphics[width=0.21\linewidth]{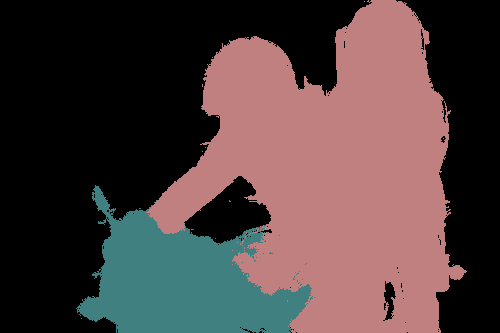} &
    \includegraphics[width=0.21\linewidth]{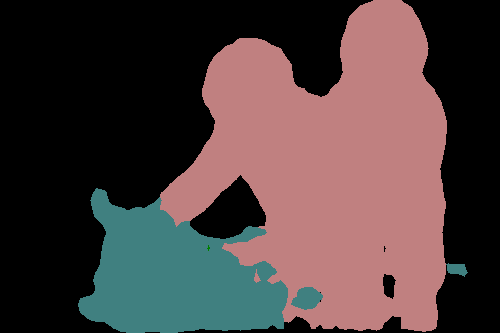} &
    \includegraphics[width=0.21\linewidth]{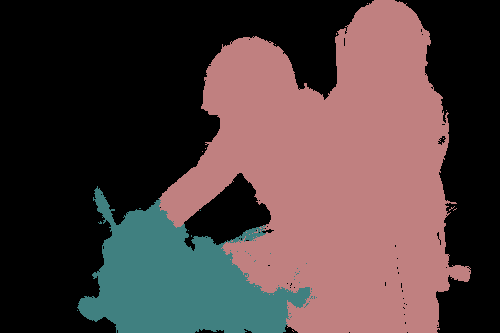} \\

    {\scriptsize Image} &
    {\scriptsize VGG-16 Bef.} &
    {\scriptsize VGG-16 Aft.} &
    {\scriptsize ResNet Bef.} &
    {\scriptsize ResNet Aft.} \\
  \end{tabular}
  }
  \caption{DeepLab results based on VGG-16 net or ResNet-101 before and after CRF.
    The CRF is critical for accurate prediction along object boundaries with VGG-16, whereas
    ResNet-101 has acceptable performance even before CRF.}
  \label{fig:res_vs_vgg_results}
\end{figure}

\begin{figure}[!t]
\centering
\resizebox{\columnwidth}{!}{
  \begin{tabular} {c c}
    \raisebox{1cm} {
    \begin{tabular}{c c}
      \includegraphics[height=0.1\linewidth]{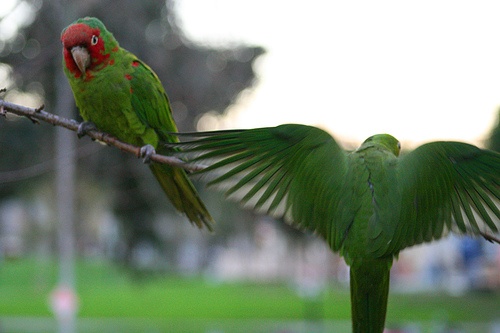} &
      \includegraphics[height=0.1\linewidth]{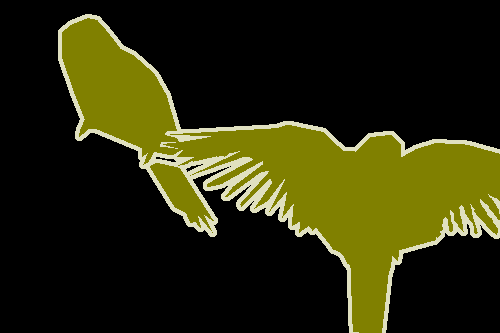} \\
      \includegraphics[height=0.1\linewidth]{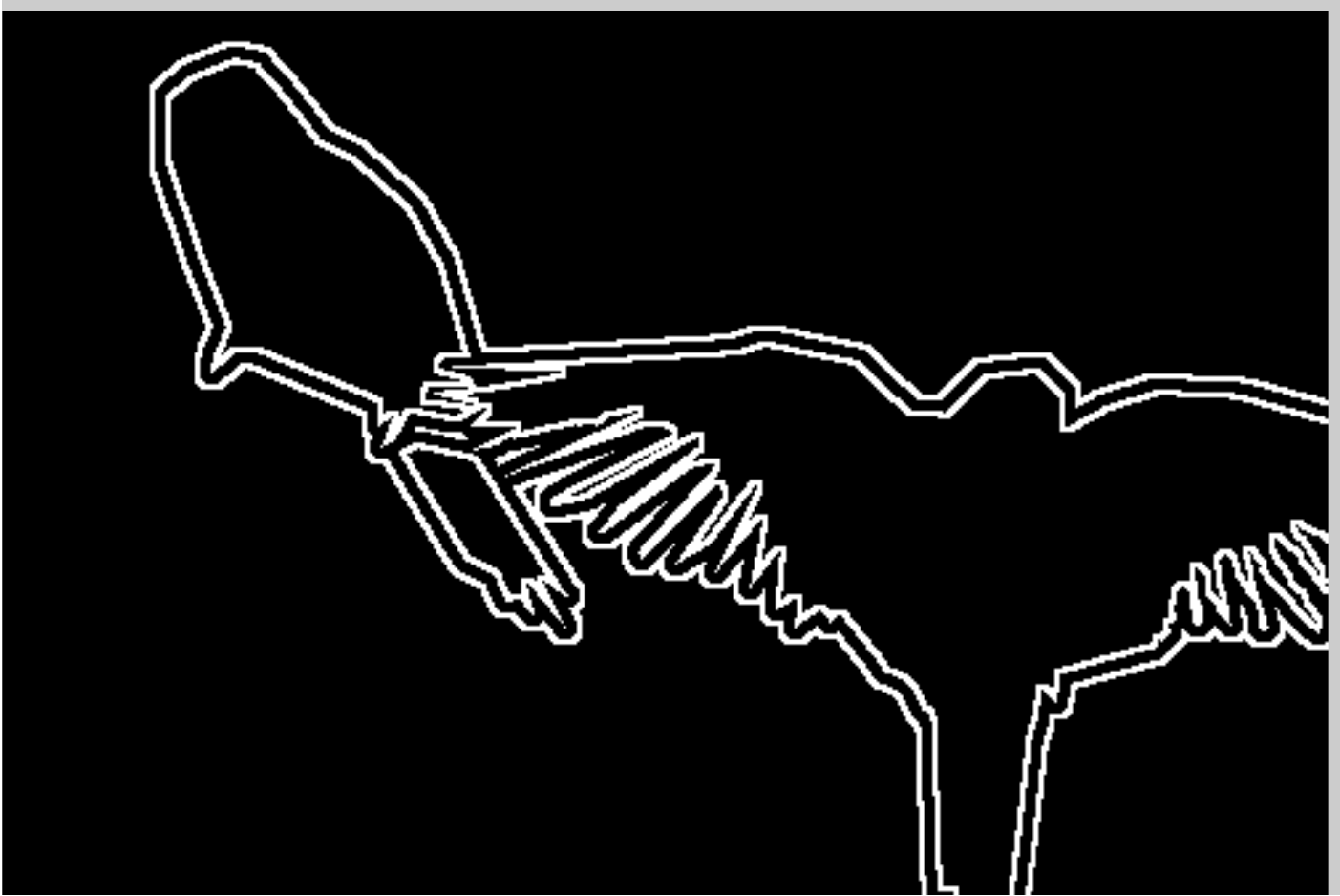} &
      \includegraphics[height=0.1\linewidth]{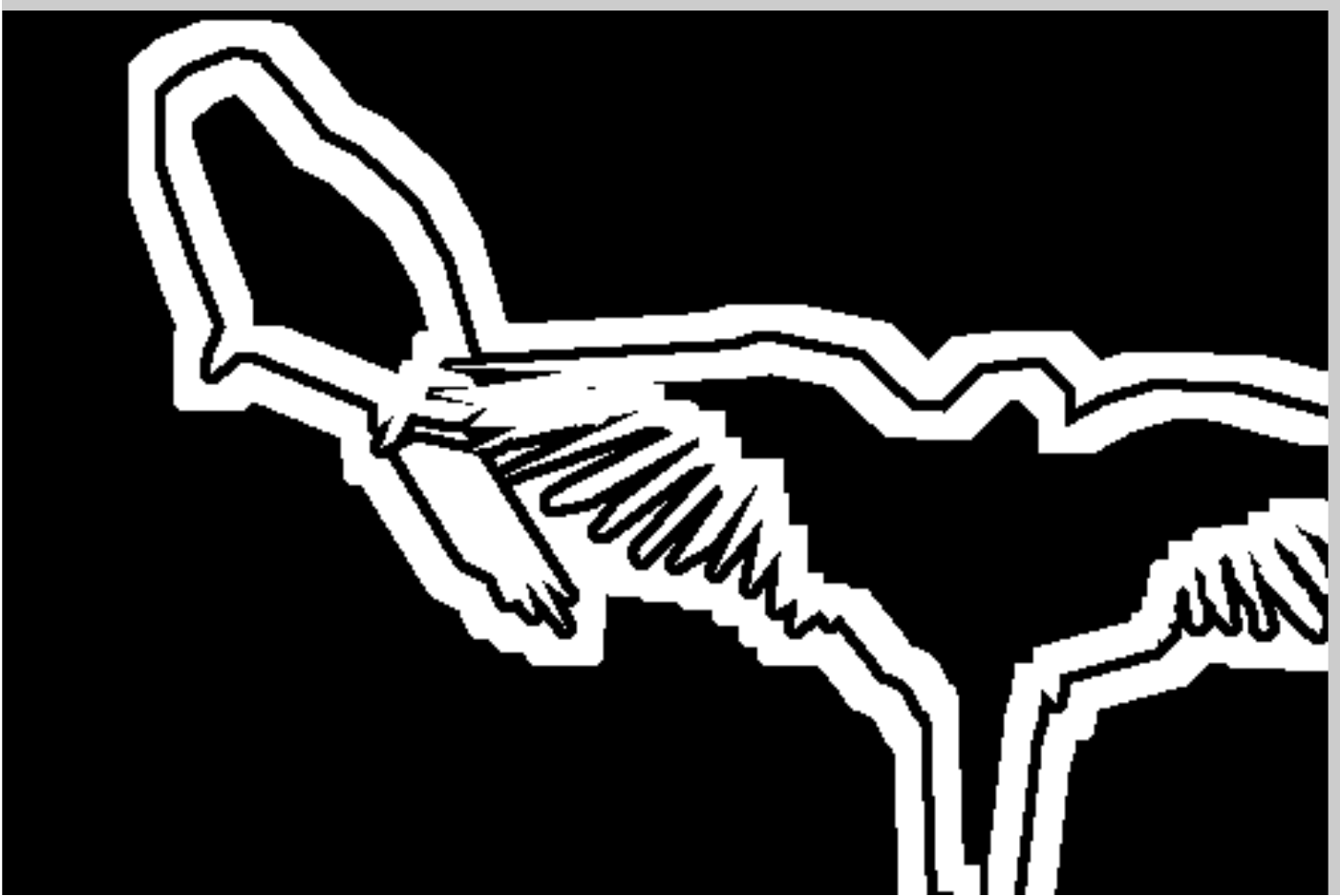} \\
    \end{tabular} } &
    \includegraphics[height=0.25\linewidth]{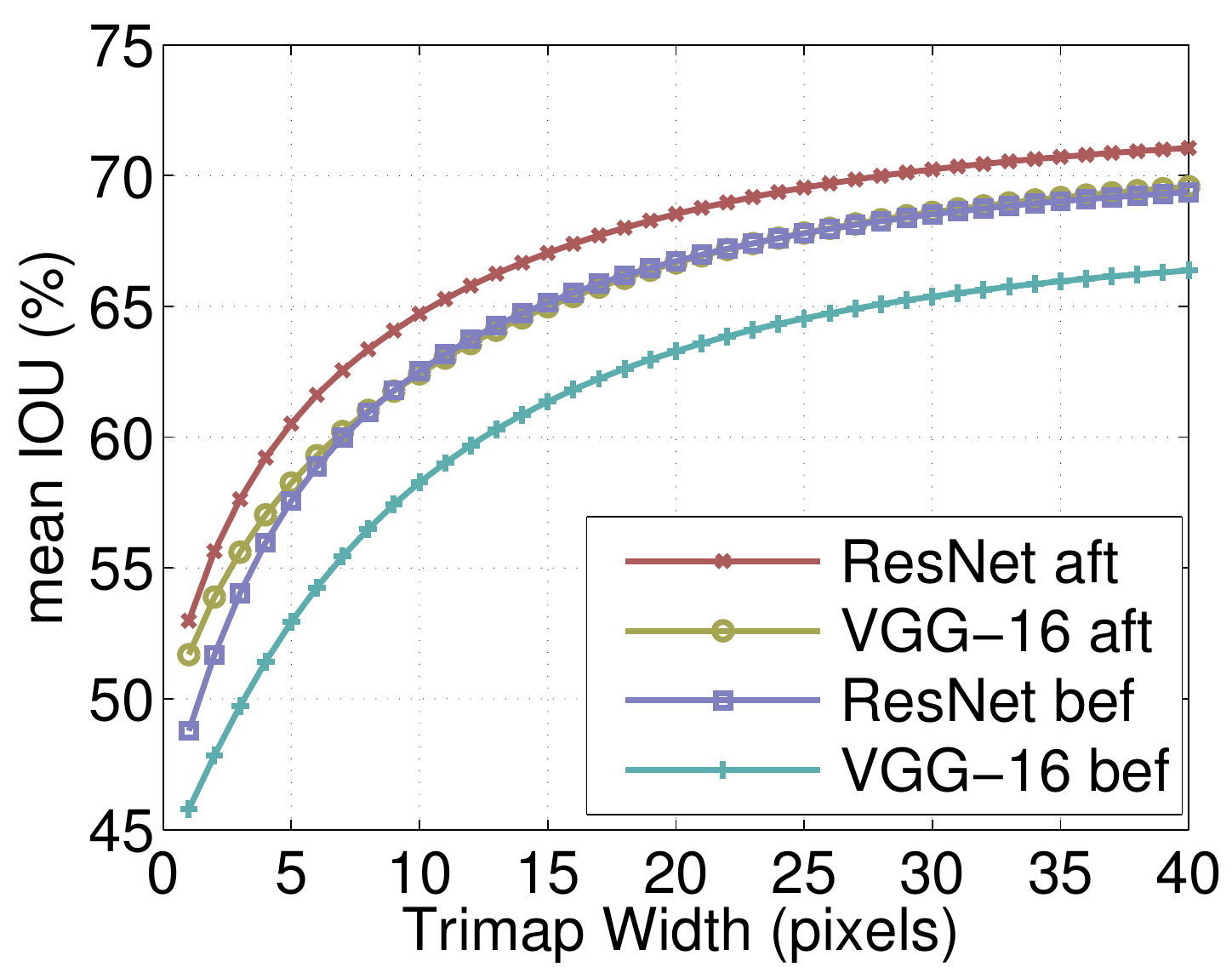} \\
    (a) & (b) \\
   \end{tabular}
}
  \caption{(a) Trimap examples (top-left: image. top-right: ground-truth. bottom-left: trimap of 2 pixels.
    bottom-right: trimap of 10 pixels). (b) Pixel mean IOU as a function of the band width around the
    object boundaries when employing VGG-16 or ResNet-101 before and after CRF.}
  \label{fig:IOUBoundary}
\end{figure}

\textbf{VGG-16 \vs ResNet-101:} We have observed that DeepLab based on ResNet-101 \cite{he2015deep}
delivers better segmentation results along object boundaries than employing VGG-16 \cite{simonyan2014very}, as
visualized in \figref{fig:res_vs_vgg_results}. We think the identity mapping \cite{he2016identity} of ResNet-101
has similar effect as hyper-column features \cite{hariharan2014hypercolumns}, which exploits the features from
the intermediate layers to better localize boundaries. We further quantize this effect in \figref{fig:IOUBoundary} within the
``trimap'' \cite{kohli2009robust, krahenbuhl2011efficient} (a narrow band along object boundaries). As shown in
the figure, employing ResNet-101 before CRF has almost the same accuracy along object boundaries as employing
VGG-16 in conjunction with a CRF. Post-processing the ResNet-101 result with a CRF further improves the segmentation
result.

\subsection{PASCAL-Context}
\label{exp:pascal_context}

\begin{figure*}[!t]
  \centering
  \scalebox{0.9} {
  \begin{tabular}{c}
    \includegraphics[width=1\linewidth]{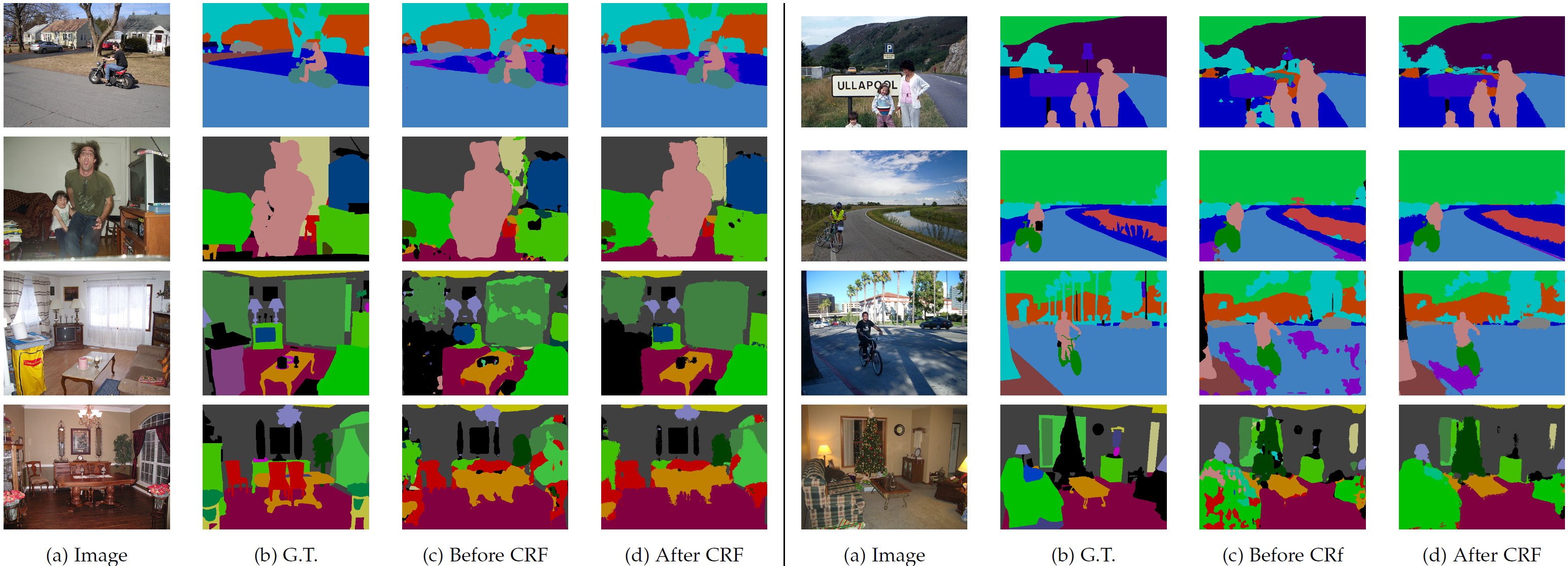} \\
  \end{tabular}
  }
  \caption{PASCAL-Context results. Input image, ground-truth,
    and our DeepLab results before/after CRF.}
  \label{fig:pascal_context_val_results}
\end{figure*}

\begin{table}[!t]
  \centering
  \addtolength{\tabcolsep}{-3pt}
  \begin{tabular} {l c c c c c c | c}
    \toprule[0.2 em]
    {\bf Method} & {\bf MSC} & {\bf COCO} & {\bf Aug} & {\bf LargeFOV} & {\bf ASPP} & {\bf CRF} & {\bf mIOU} \\
    \toprule[0.2 em]
    \multicolumn{7}{l}{\it VGG-16} & \\
    DeepLab \cite{chen2014semantic}& & & &\checkmark & & & 37.6 \\
    DeepLab \cite{chen2014semantic}& & & &\checkmark & & \checkmark  &  39.6 \\
    \midrule
    \multicolumn{7}{l}{\it ResNet-101} & \\
    DeepLab & & & & & & &  39.6 \\
    DeepLab &\checkmark & & \checkmark & & & &  41.4 \\
    DeepLab &\checkmark &\checkmark & \checkmark & & & &  42.9 \\
    DeepLab &\checkmark &\checkmark & \checkmark & \checkmark & & & 43.5 \\
    DeepLab &\checkmark &\checkmark & \checkmark & & \checkmark & & 44.7 \\
    DeepLab &\checkmark &\checkmark & \checkmark & & \checkmark & \checkmark & 45.7 \\
    \midrule \midrule
    $O_2P$ \cite{carreira2012semantic}& & & & & &  & 18.1 \\
    CFM \cite{dai2014convolutional}& & & & & &  & 34.4 \\
    FCN-8s \cite{long2014fully}& & & & & &  & 37.8 \\
    CRF-RNN \cite{zheng2015conditional}& & & & & &  & 39.3 \\
    ParseNet \cite{liu2015parsenet}& & & & & &  & 40.4 \\
    BoxSup \cite{dai2015boxsup}& & & & & &  & 40.5 \\
    HO\_CRF \cite{arnab2015higher}& & & & & &  & 41.3 \\
    Context \cite{lin2015efficient}& & & & & &  & 43.3 \\
    VeryDeep \cite{wu2016bridging}& & & & & &  & 44.5 \\
    \bottomrule[0.1 em]
  \end{tabular}
  \caption{Comparison with other state-of-art methods on PASCAL-Context dataset.}
  \label{tab:pascal_context}
\end{table}

\textbf{Dataset:} The PASCAL-Context dataset \cite{mottaghi2014role} provides
detailed semantic labels for the whole scene, including both object (\eg, person)
and stuff (\eg, sky). Following \cite{mottaghi2014role}, the proposed models are
evaluated on the most frequent 59 classes along with one background category.
The training set and validation set contain 4998 and 5105 images.

\textbf{Evaluation:} We report the evaluation results in \tabref{tab:pascal_context}.
Our VGG-16 based LargeFOV variant yields 37.6\% before and 39.6\% after CRF.
Repurposing the ResNet-101 \cite{he2015deep} for DeepLab improves 2\% over the
VGG-16 LargeFOV. Similar to \cite{chen2015attention}, employing multi-scale inputs
and max-pooling to merge the results improves the performance to 41.4\%.
Pretraining the model on MS-COCO brings extra 1.5\% improvement.
Employing atrous spatial pyramid pooling is more effective than LargeFOV.
After further employing dense CRF as post processing, our final model
yields 45.7\%, outperforming the current state-of-art method
\cite{lin2015efficient} by 2.4\% without using their non-linear pairwise term. Our final
model is slightly better than the concurrent work \cite{wu2016bridging} by 1.2\%, which also employs
atrous convolution to repurpose the residual net of \cite{he2015deep} for semantic segmentation.

\textbf{Qualitative results:} We visualize the segmentation results of our best
model with and without CRF as post processing in
\figref{fig:pascal_context_val_results}. DeepLab before CRF can already predict
most of the object/stuff with high accuracy. Employing CRF, our model is able to
further remove isolated false positives and improve the prediction along
object/stuff boundaries.

\subsection{PASCAL-Person-Part}
\label{exp:pascal_person_part}

\begin{figure*}[!th]
  \centering
  \scalebox{0.9} {
  \begin{tabular}{c}
    \includegraphics[width=1.\linewidth]{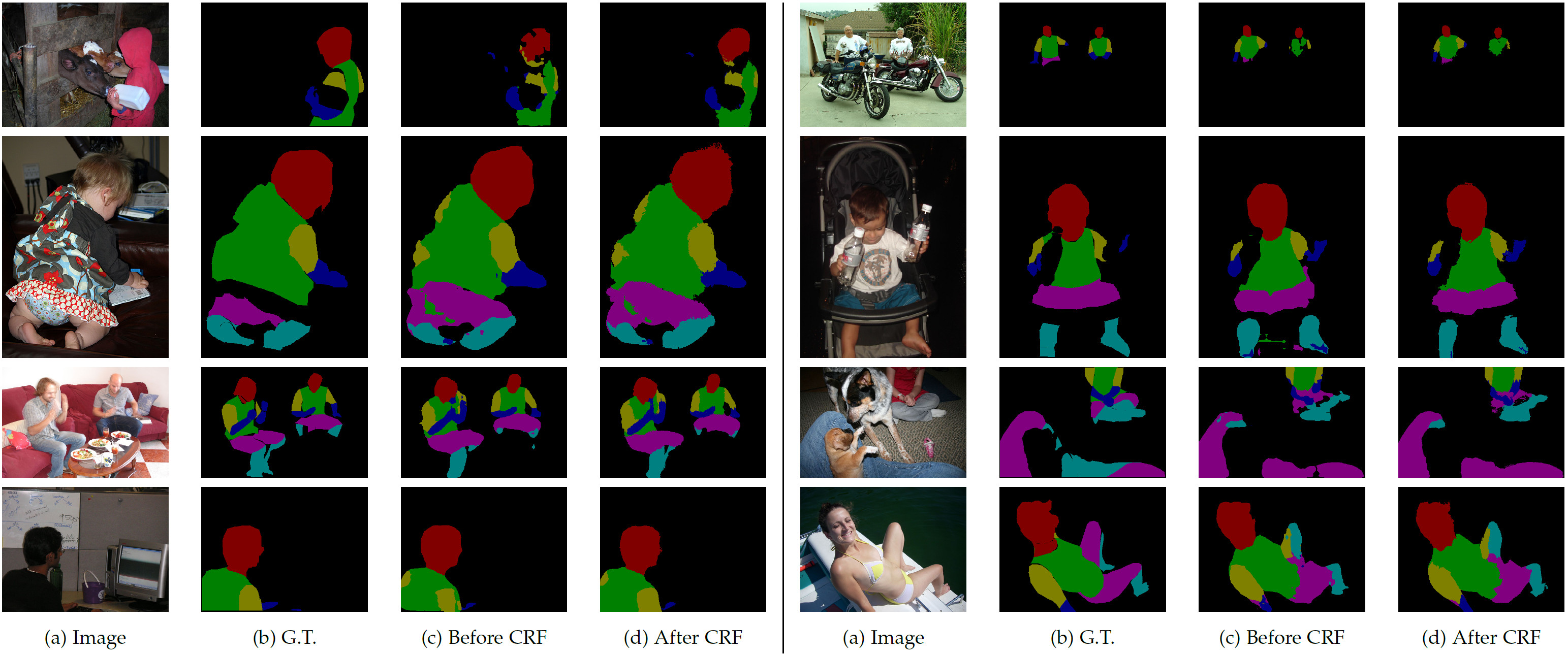} \\
  \end{tabular}
  }
  \caption{PASCAL-Person-Part results. Input image, ground-truth,
    and our DeepLab results before/after CRF.}
  \label{fig:voc10_part_val_results}
\end{figure*}

\begin{table}[!t]
  \centering
  \addtolength{\tabcolsep}{-3pt}
  \begin{tabular} {l  c c c c c c | c}
    \toprule[0.2 em]
    {\bf Method} & {\bf MSC} & {\bf COCO} & {\bf Aug} & {\bf LFOV} & {\bf ASPP} & {\bf CRF} & {\bf mIOU} \\
    \toprule[0.2 em]
    \multicolumn{7}{l}{\it ResNet-101} & \\
    DeepLab      & & & & & & & 58.90 \\
    DeepLab      & \checkmark & & \checkmark & & & & 63.10 \\
    DeepLab      & \checkmark & \checkmark & \checkmark & & & & 64.40 \\
    DeepLab      & \checkmark & \checkmark & \checkmark & & & \checkmark & 64.94 \\
    \midrule
    DeepLab      & \checkmark & \checkmark & \checkmark & \checkmark & & & 62.18 \\
    DeepLab      & \checkmark & \checkmark & \checkmark & & \checkmark & & 62.76 \\
    \midrule \midrule
    Attention \cite{chen2015attention} & & & & & & & 56.39 \\
    HAZN \cite{xia2015zoom} & & & & & &  & 57.54 \\
    LG-LSTM \cite{liang2015semantic} & & & & & &  & 57.97 \\
    Graph LSTM \cite{liang2016semantic} & & & & & &  & 60.16 \\
    \bottomrule[0.1 em]
  \end{tabular}
  \caption{Comparison with other state-of-art methods on PASCAL-Person-Part dataset.}
  \label{tab:pascal_person_part}
\end{table}

\textbf{Dataset:} We further perform experiments on semantic part segmentation \cite{wang2014semantic, wang2015joint},
using the extra PASCAL VOC 2010 annotations by \cite{chen_cvpr14}. We focus on the
{\it person} part for the dataset, which contains more training data and large
variation in object scale and human pose. Specifically, the dataset contains
detailed part annotations for every person, \eg eyes, nose. We merge the
annotations to be Head, Torso, Upper/Lower Arms and Upper/Lower Legs, resulting
in six person part classes and one background class. We only use those images
containing persons for training (1716 images) and validation (1817 images).

\textbf{Evaluation:} The human part segmentation results on PASCAL-Person-Part is
reported in \tabref{tab:pascal_person_part}. \cite{chen2015attention} has already
conducted experiments on this dataset with re-purposed VGG-16 net for DeepLab,
attaining 56.39\% (with multi-scale inputs). Therefore, in this part, we mainly
focus on the effect of repurposing ResNet-101 for DeepLab. With ResNet-101,
DeepLab alone yields 58.9\%, significantly outperforming DeepLab-LargeFOV
(VGG-16 net) and DeepLab-Attention (VGG-16 net) by about 7\% and 2.5\%,
respectively. Incorporating multi-scale inputs and fusion by max-pooling
further improves performance to 63.1\%. Additionally pretraining the model on
MS-COCO yields another 1.3\% improvement. However, we do not observe any
improvement when adopting either LargeFOV or ASPP on this dataset. Employing
the dense CRF to post process our final output substantially outperforms the
concurrent work \cite{liang2016semantic} by 4.78\%.

\textbf{Qualitative results:} We visualize the results in \figref{fig:voc10_part_val_results}.

\subsection{Cityscapes}
\label{exp:cityscapes}

\begin{figure*}[!t]
  \centering
  \scalebox{1} {
  \begin{tabular}{c}
    \includegraphics[width=0.9\linewidth]{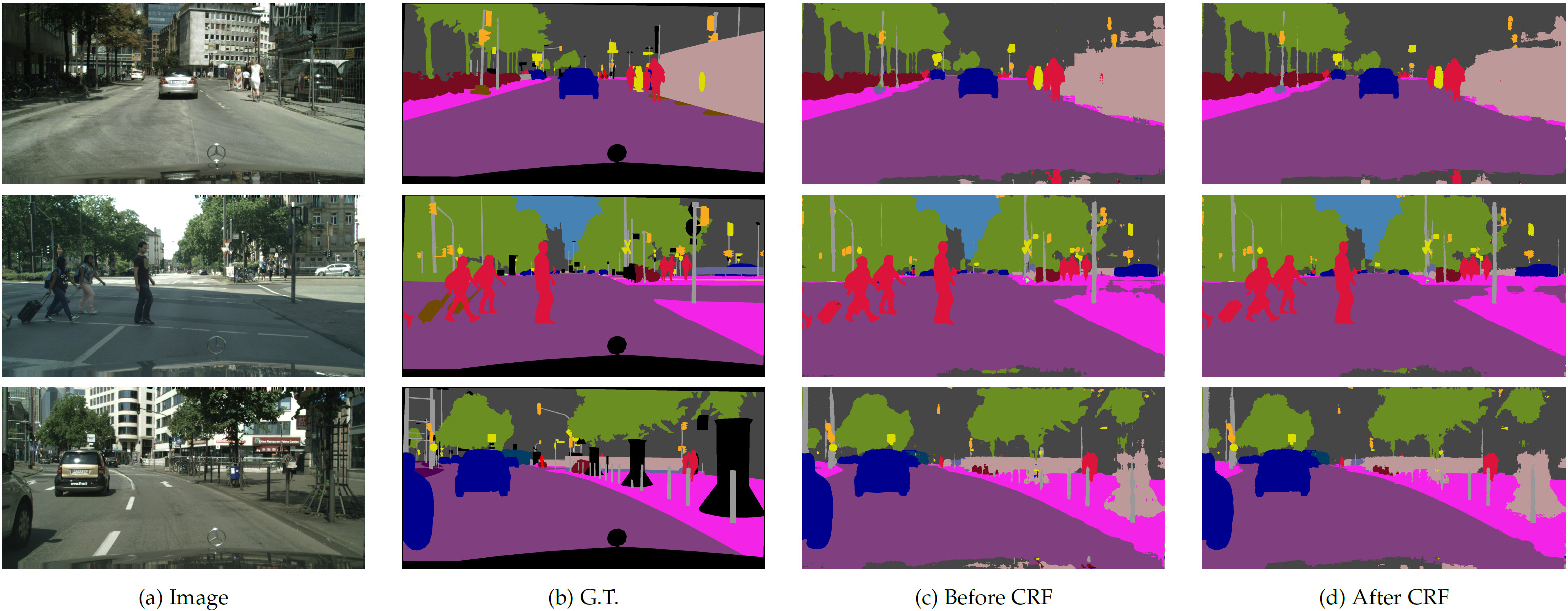} \\
  \end{tabular}
  }
  \caption{Cityscapes results. Input image, ground-truth,
    and our DeepLab results before/after CRF.}
  \label{fig:cityscapes_val_results}
\end{figure*}

\begin{table}[!t]
  \centering
  \addtolength{\tabcolsep}{2.5pt}
  \begin{tabular}{l | c}
    \toprule[0.2 em]
    {\bf Method} & {\bf mIOU} \\
    \toprule[0.2 em]
    \multicolumn{2}{l}{\it pre-release version of dataset} \\
    Adelaide\_Context \cite{lin2015efficient} & 66.4 \\
    FCN-8s \cite{long2014fully} & 65.3 \\
    \midrule
    DeepLab-CRF-LargeFOV-StrongWeak \cite{papandreou2015weakly} & 64.8 \\
    DeepLab-CRF-LargeFOV \cite{chen2014semantic} & 63.1 \\
    \midrule
    CRF-RNN \cite{zheng2015conditional} & 62.5 \\
    DPN \cite{liu2015semantic} & 59.1 \\
    Segnet basic \cite{badrinarayanan2015segnet} & 57.0 \\
    Segnet extended \cite{badrinarayanan2015segnet} & 56.1 \\
    \midrule \midrule
    \multicolumn{2}{l}{\it official version} \\
    Adelaide\_Context \cite{lin2015efficient} & 71.6 \\
    Dilation10 \cite{yu2015multi} & 67.1 \\
    DPN \cite{liu2015semantic} & 66.8  \\
    Pixel-level Encoding \cite{uhrig2016pixel} & 64.3 \\
    \midrule
    DeepLab-CRF (ResNet-101) & 70.4 \\
    \bottomrule[0.1 em]
  \end{tabular}
  \caption{Test set results on the Cityscapes dataset, comparing our DeepLab system with other state-of-art methods.}
  \label{tab:cityscapes_test}
\end{table}

\begin{table}[!t]
  \centering
  \addtolength{\tabcolsep}{2pt}
  \begin{tabular}{c c c c c | c}
    \toprule[0.2 em]
    {\bf Full} & {\bf Aug} & {\bf LargeFOV} & {\bf ASPP} & {\bf CRF} & {\bf mIOU} \\
    \toprule[0.2 em]
    \multicolumn{5}{l}{\it VGG-16} & \\
    & & \checkmark & & & 62.97 \\
    & & \checkmark & & \checkmark & 64.18 \\
    \checkmark & & \checkmark & & & 64.89 \\
    \checkmark & & \checkmark & & \checkmark & 65.94 \\
    \midrule
    \multicolumn{5}{l}{\it ResNet-101} & \\
    \checkmark & & & & & 66.6 \\
    \checkmark & & \checkmark & & & 69.2 \\
    \checkmark & &  & \checkmark & & 70.4 \\
    \checkmark & \checkmark & & \checkmark & & 71.0 \\
    \checkmark & \checkmark & & \checkmark & \checkmark & 71.4 \\
    \midrule
    \bottomrule[0.1 em]
  \end{tabular}
  \caption{Val set results on Cityscapes dataset. {\bf Full}: model trained with full resolution images.}
  \label{tab:cityscapes_val_resnet}
\end{table}

\textbf{Dataset:} Cityscapes \cite{Cordts2016Cityscapes} is a recently released
large-scale dataset, which contains high quality pixel-level annotations of 5000
images collected in street scenes from 50 different cities. Following the
evaluation protocol \cite{Cordts2016Cityscapes}, 19 semantic labels (belonging
to 7 super categories: ground, construction, object, nature, sky, human, and
vehicle) are used for evaluation (the void label is not considered for
evaluation). The training, validation, and test sets contain 2975, 500, and
1525 images respectively.

\textbf{Test set results of pre-release:} We have participated in benchmarking
the Cityscapes dataset pre-release. As shown in the top of \tabref{tab:cityscapes_test},
our model attained third place, with performance of 63.1\% and 64.8\% (with training on
additional coarsely annotated images). 

\textbf{Val set results:} After the initial release, we further explored the
validation set in \tabref{tab:cityscapes_val_resnet}. The images of Cityscapes
have resolution \by{2048}{1024}, making it a challenging problem to train deeper
networks with limited GPU memory. During benchmarking the pre-release of the dataset,
we downsampled the images by 2. However, we have found that it is beneficial to
process the images in their original resolution. With the same training protocol,
using images of original resolution significantly brings 1.9\% and 1.8\% improvements
before and after CRF, respectively. In order to perform inference on this dataset with
high resolution images, we split each image into overlapped regions, similar to
\cite{Cordts2016Cityscapes}. We have also replaced the VGG-16 net with ResNet-101.
We do not exploit multi-scale inputs due to the limited GPU memories at hand.
Instead, we only explore (1) deeper networks (\ie, ResNet-101), (2) data augmentation,
(3) LargeFOV or ASPP, and (4) CRF as post processing on this dataset. We first find
that employing ResNet-101 alone is better than using VGG-16 net. Employing LargeFOV
brings 2.6\% improvement and using ASPP further improves results by 1.2\%.
Adopting data augmentation and CRF as post processing brings another 0.6\% and 0.4\%,
respectively.

\textbf{Current test result:} We have uploaded our best model to the evaluation server,
obtaining performance of 70.4\%. Note that our model is only trained on the train set.

\textbf{Qualitative results:} We visualize the results in \figref{fig:cityscapes_val_results}.

\subsection{Failure Modes}
\begin{figure}[!t]
\centering
\resizebox{\columnwidth}{!}{
  \begin{tabular} {c c c c}
    \includegraphics[width=0.24\linewidth]{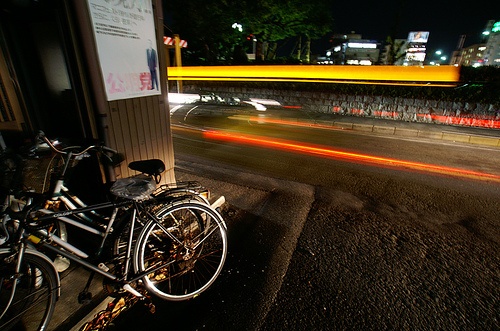} &
    \includegraphics[width=0.24\linewidth]{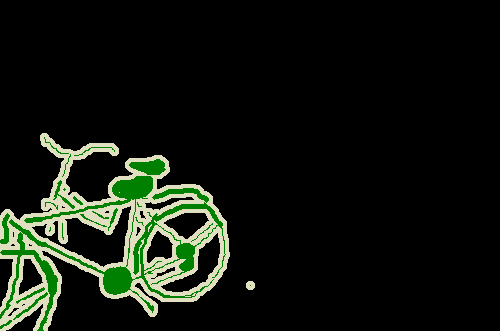} &
    \includegraphics[width=0.24\linewidth]{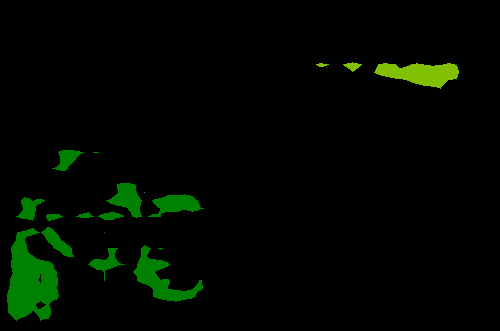} &
    \includegraphics[width=0.24\linewidth]{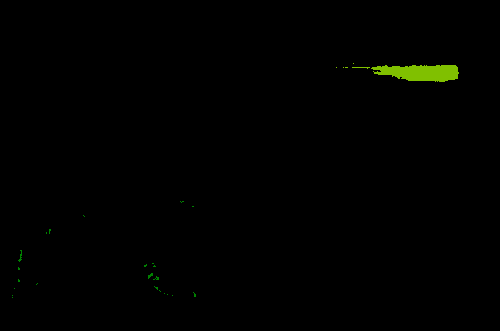} \\
    \includegraphics[width=0.24\linewidth]{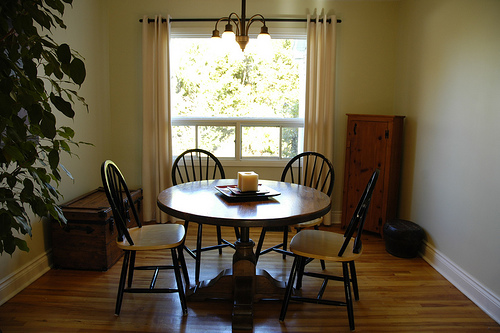} &
    \includegraphics[width=0.24\linewidth]{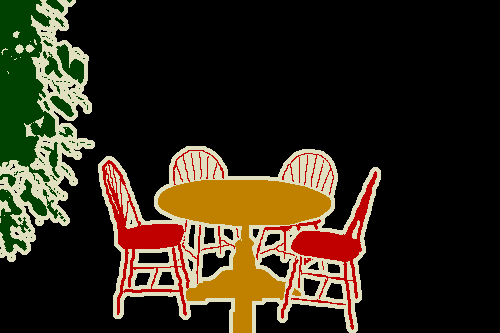} &
    \includegraphics[width=0.24\linewidth]{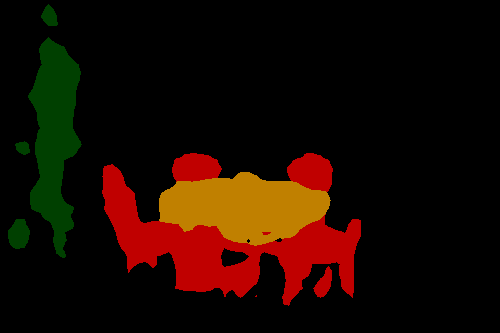} &
    \includegraphics[width=0.24\linewidth]{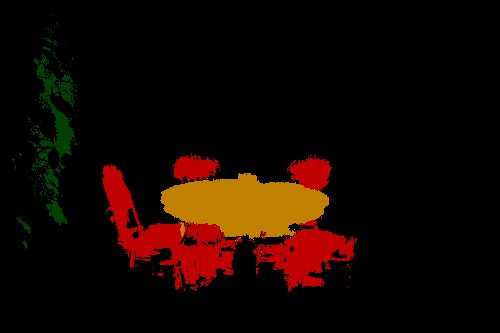} \\
                    {\scriptsize (a) Image} &
                    {\scriptsize (b) G.T.} &
                    {\scriptsize (c) Before CRF} &
                    {\scriptsize (d) After CRF} \\
  \end{tabular}

}
  \caption{Failure modes. Input image, ground-truth, and our DeepLab results before/after CRF.}  
  \label{fig:failure_modes}
\end{figure}

We further qualitatively analyze some failure modes of our best model variant on PASCAL VOC 2012 {\it val} set. As shown in \figref{fig:failure_modes}, our proposed model fails to capture the delicate boundaries of objects, such as bicycle and chair. The details could not even be recovered by the CRF post processing since the unary term is not confident enough. We hypothesize the encoder-decoder structure of 
\cite{badrinarayanan2015segnet, ronneberger2015u} may alleviate the problem by exploiting the high resolution feature maps in the decoder path. How to efficiently incorporate the method is left as a future work.

\section{Conclusion}
Our proposed ``DeepLab'' system re-purposes networks trained
on image classification to the task of semantic segmentation by applying the `atrous convolution' with upsampled filters for dense feature extraction. We further extend it to atrous spatial pyramid pooling, which encodes objects as well as image context at multiple scales. To produce semantically accurate predictions and detailed segmentation maps along object boundaries, we also combine ideas from deep convolutional neural networks and fully-connected conditional random fields. Our experimental results show that the
proposed method significantly advances the state-of-art in several challenging
datasets, including PASCAL VOC 2012 semantic image segmentation benchmark, PASCAL-Context, PASCAL-Person-Part, and Cityscapes datasets.

\section*{Acknowledgments}
This work was partly supported by the ARO 62250-CS, FP7-RECONFIG, FP7-MOBOT, and H2020-ISUPPORT EU projects. We gratefully acknowledge the support of NVIDIA Corporation with the donation of GPUs used for this research.

\ifCLASSOPTIONcaptionsoff
  \newpage
\fi



%

{
\bibliographystyle{IEEEtran}
\bibliography{./egbib}

\begin{thebibliography}{100}
\providecommand{\url}[1]{#1}
\csname url@samestyle\endcsname
\providecommand{\newblock}{\relax}
\providecommand{\bibinfo}[2]{#2}
\providecommand{\BIBentrySTDinterwordspacing}{\spaceskip=0pt\relax}
\providecommand{\BIBentryALTinterwordstretchfactor}{4}
\providecommand{\BIBentryALTinterwordspacing}{\spaceskip=\fontdimen2\font plus
\BIBentryALTinterwordstretchfactor\fontdimen3\font minus
  \fontdimen4\font\relax}
\providecommand{\BIBforeignlanguage}[2]{{%
\expandafter\ifx\csname l@#1\endcsname\relax
\typeout{** WARNING: IEEEtran.bst: No hyphenation pattern has been}%
\typeout{** loaded for the language `#1'. Using the pattern for}%
\typeout{** the default language instead.}%
\else
\language=\csname l@#1\endcsname
\fi
#2}}
\providecommand{\BIBdecl}{\relax}
\BIBdecl

\bibitem{LeCun1998}
Y.~LeCun, L.~Bottou, Y.~Bengio, and P.~Haffner, ``{Gradient-based learning
  applied to document recognition},'' in \emph{Proc. IEEE}, 1998.

\bibitem{KrizhevskyNIPS2013}
A.~Krizhevsky, I.~Sutskever, and G.~E. Hinton, ``Imagenet classification with
  deep convolutional neural networks,'' in \emph{NIPS}, 2013.

\bibitem{sermanet2013overfeat}
P.~Sermanet, D.~Eigen, X.~Zhang, M.~Mathieu, R.~Fergus, and Y.~LeCun,
  ``Overfeat: Integrated recognition, localization and detection using
  convolutional networks,'' \emph{arXiv:1312.6229}, 2013.

\bibitem{simonyan2014very}
K.~Simonyan and A.~Zisserman, ``Very deep convolutional networks for
  large-scale image recognition,'' in \emph{ICLR}, 2015.

\bibitem{szegedy2014going}
C.~Szegedy, W.~Liu, Y.~Jia, P.~Sermanet, S.~Reed, D.~Anguelov, D.~Erhan,
  V.~Vanhoucke, and A.~Rabinovich, ``Going deeper with convolutions,''
  \emph{arXiv:1409.4842}, 2014.

\bibitem{papandreou2014untangling}
G.~Papandreou, I.~Kokkinos, and P.-A. Savalle, ``Modeling local and global
  deformations in deep learning: Epitomic convolution, multiple instance
  learning, and sliding window detection,'' in \emph{CVPR}, 2015.

\bibitem{girshick2014rcnn}
R.~Girshick, J.~Donahue, T.~Darrell, and J.~Malik, ``Rich feature hierarchies
  for accurate object detection and semantic segmentation,'' in \emph{CVPR},
  2014.

\bibitem{erhan2014scalable}
D.~Erhan, C.~Szegedy, A.~Toshev, and D.~Anguelov, ``Scalable object detection
  using deep neural networks,'' in \emph{CVPR}, 2014.

\bibitem{girshick2015fast}
R.~Girshick, ``Fast r-cnn,'' in \emph{ICCV}, 2015.

\bibitem{ren2015faster}
S.~Ren, K.~He, R.~Girshick, and J.~Sun, ``Faster r-cnn: Towards real-time
  object detection with region proposal networks,'' in \emph{NIPS}, 2015.

\bibitem{he2015deep}
K.~He, X.~Zhang, S.~Ren, and J.~Sun, ``Deep residual learning for image
  recognition,'' \emph{arXiv:1512.03385}, 2015.

\bibitem{liu2015ssd}
W.~Liu, D.~Anguelov, D.~Erhan, C.~Szegedy, and S.~Reed, ``{SSD}: Single shot
  multibox detector,'' \emph{arXiv:1512.02325}, 2015.

\bibitem{zeiler2014visualizing}
M.~D. Zeiler and R.~Fergus, ``Visualizing and understanding convolutional
  networks,'' in \emph{ECCV}, 2014.

\bibitem{long2014fully}
J.~Long, E.~Shelhamer, and T.~Darrell, ``Fully convolutional networks for
  semantic segmentation,'' in \emph{CVPR}, 2015.

\bibitem{holschneider1989real}
M.~Holschneider, R.~Kronland-Martinet, J.~Morlet, and P.~Tchamitchian, ``A
  real-time algorithm for signal analysis with the help of the wavelet
  transform,'' in \emph{Wavelets: Time-Frequency Methods and Phase Space},
  1989, pp. 289--297.

\bibitem{giusti2013fast}
A.~Giusti, D.~Ciresan, J.~Masci, L.~Gambardella, and J.~Schmidhuber, ``Fast
  image scanning with deep max-pooling convolutional neural networks,'' in
  \emph{ICIP}, 2013.

\bibitem{chen2015attention}
L.-C. Chen, Y.~Yang, J.~Wang, W.~Xu, and A.~L. Yuille, ``Attention to scale:
  Scale-aware semantic image segmentation,'' in \emph{CVPR}, 2016.

\bibitem{kokkinos2016pushing}
I.~Kokkinos, ``Pushing the boundaries of boundary detection using deep
  learning,'' in \emph{ICLR}, 2016.

\bibitem{lazebnik2006beyond}
S.~Lazebnik, C.~Schmid, and J.~Ponce, ``Beyond bags of features: Spatial
  pyramid matching for recognizing natural scene categories,'' in \emph{CVPR},
  2006.

\bibitem{he2014spatial}
K.~He, X.~Zhang, S.~Ren, and J.~Sun, ``Spatial pyramid pooling in deep
  convolutional networks for visual recognition,'' in \emph{ECCV}, 2014.

\bibitem{hariharan2014hypercolumns}
B.~Hariharan, P.~Arbel{\'a}ez, R.~Girshick, and J.~Malik, ``Hypercolumns for
  object segmentation and fine-grained localization,'' in \emph{CVPR}, 2015.

\bibitem{krahenbuhl2011efficient}
P.~Kr{\"a}henb{\"u}hl and V.~Koltun, ``Efficient inference in fully connected
  crfs with gaussian edge potentials,'' in \emph{NIPS}, 2011.

\bibitem{rother2004grabcut}
C.~Rother, V.~Kolmogorov, and A.~Blake, ``{GrabCut}: Interactive foreground
  extraction using iterated graph cuts,'' in \emph{SIGGRAPH}, 2004.

\bibitem{shotton2009textonboost}
J.~Shotton, J.~Winn, C.~Rother, and A.~Criminisi, ``Textonboost for image
  understanding: Multi-class object recognition and segmentation by jointly
  modeling texture, layout, and context,'' \emph{IJCV}, 2009.

\bibitem{lucchi2011spatial}
A.~Lucchi, Y.~Li, X.~Boix, K.~Smith, and P.~Fua, ``Are spatial and global
  constraints really necessary for segmentation?'' in \emph{ICCV}, 2011.

\bibitem{he2004multiscale}
X.~He, R.~S. Zemel, and M.~Carreira-Perpindn, ``Multiscale conditional random
  fields for image labeling,'' in \emph{CVPR}, 2004.

\bibitem{ladicky2009associative}
L.~Ladicky, C.~Russell, P.~Kohli, and P.~H. Torr, ``Associative hierarchical
  crfs for object class image segmentation,'' in \emph{ICCV}, 2009.

\bibitem{lempitsky2011pylon}
V.~Lempitsky, A.~Vedaldi, and A.~Zisserman, ``Pylon model for semantic
  segmentation,'' in \emph{NIPS}, 2011.

\bibitem{delong2012fast}
A.~Delong, A.~Osokin, H.~N. Isack, and Y.~Boykov, ``Fast approximate energy
  minimization with label costs,'' \emph{IJCV}, 2012.

\bibitem{gonfaus2010harmony}
J.~M. Gonfaus, X.~Boix, J.~Van~de Weijer, A.~D. Bagdanov, J.~Serrat, and
  J.~Gonzalez, ``Harmony potentials for joint classification and
  segmentation,'' in \emph{CVPR}, 2010.

\bibitem{kohli2009robust}
P.~Kohli, P.~H. Torr \emph{et~al.}, ``Robust higher order potentials for
  enforcing label consistency,'' \emph{IJCV}, vol.~82, no.~3, pp. 302--324,
  2009.

\bibitem{CPY13}
L.-C. Chen, G.~Papandreou, and A.~Yuille, ``Learning a dictionary of shape
  epitomes with applications to image labeling,'' in \emph{ICCV}, 2013.

\bibitem{Wang15}
P.~Wang, X.~Shen, Z.~Lin, S.~Cohen, B.~Price, and A.~Yuille, ``Towards unified
  depth and semantic prediction from a single image,'' in \emph{CVPR}, 2015.

\bibitem{everingham2014pascal}
M.~Everingham, S.~M.~A. Eslami, L.~V. Gool, C.~K.~I. Williams, J.~Winn, and
  A.~Zisserma, ``The pascal visual object classes challenge – a
  retrospective,'' \emph{IJCV}, 2014.

\bibitem{mottaghi2014role}
R.~Mottaghi, X.~Chen, X.~Liu, N.-G. Cho, S.-W. Lee, S.~Fidler, R.~Urtasun, and
  A.~Yuille, ``The role of context for object detection and semantic
  segmentation in the wild,'' in \emph{CVPR}, 2014.

\bibitem{chen_cvpr14}
X.~Chen, R.~Mottaghi, X.~Liu, S.~Fidler, R.~Urtasun, and A.~Yuille, ``Detect
  what you can: Detecting and representing objects using holistic models and
  body parts,'' in \emph{CVPR}, 2014.

\bibitem{Cordts2016Cityscapes}
M.~Cordts, M.~Omran, S.~Ramos, T.~Rehfeld, M.~Enzweiler, R.~Benenson,
  U.~Franke, S.~Roth, and B.~Schiele, ``The cityscapes dataset for semantic
  urban scene understanding,'' in \emph{CVPR}, 2016.

\bibitem{chen2014semantic}
L.-C. Chen, G.~Papandreou, I.~Kokkinos, K.~Murphy, and A.~L. Yuille, ``Semantic
  image segmentation with deep convolutional nets and fully connected crfs,''
  in \emph{ICLR}, 2015.

\bibitem{farabet2013learning}
C.~Farabet, C.~Couprie, L.~Najman, and Y.~LeCun, ``Learning hierarchical
  features for scene labeling,'' \emph{PAMI}, 2013.

\bibitem{lin2015efficient}
G.~Lin, C.~Shen, I.~Reid \emph{et~al.}, ``Efficient piecewise training of deep
  structured models for semantic segmentation,'' \emph{arXiv:1504.01013}, 2015.

\bibitem{jia2014caffe}
Y.~Jia, E.~Shelhamer, J.~Donahue, S.~Karayev, J.~Long, R.~Girshick,
  S.~Guadarrama, and T.~Darrell, ``Caffe: Convolutional architecture for fast
  feature embedding,'' \emph{arXiv:1408.5093}, 2014.

\bibitem{TuB10}
Z.~Tu and X.~Bai, ``Auto-context and its application to high-level vision tasks
  and 3d brain image segmentation,'' \emph{{IEEE} Trans. Pattern Anal. Mach.
  Intell.}, vol.~32, no.~10, pp. 1744--1757, 2010.

\bibitem{shotton2008semantic}
J.~Shotton, M.~Johnson, and R.~Cipolla, ``Semantic texton forests for image
  categorization and segmentation,'' in \emph{CVPR}, 2008.

\bibitem{FulkersonVS09}
B.~Fulkerson, A.~Vedaldi, and S.~Soatto, ``Class segmentation and object
  localization with superpixel neighborhoods,'' in \emph{ICCV}, 2009.

\bibitem{carreira2012semantic}
J.~Carreira, R.~Caseiro, J.~Batista, and C.~Sminchisescu, ``Semantic
  segmentation with second-order pooling,'' in \emph{ECCV}, 2012.

\bibitem{carreira2012cpmc}
J.~Carreira and C.~Sminchisescu, ``{CPMC}: Automatic object segmentation using
  constrained parametric min-cuts,'' \emph{PAMI}, vol.~34, no.~7, pp.
  1312--1328, 2012.

\bibitem{arbelaez2014multiscale}
P.~Arbel{\'a}ez, J.~Pont-Tuset, J.~T. Barron, F.~Marques, and J.~Malik,
  ``Multiscale combinatorial grouping,'' in \emph{CVPR}, 2014.

\bibitem{Uijlings13}
J.~Uijlings, K.~van~de Sande, T.~Gevers, and A.~Smeulders, ``Selective search
  for object recognition,'' \emph{IJCV}, 2013.

\bibitem{hariharan2014simultaneous}
B.~Hariharan, P.~Arbel{\'a}ez, R.~Girshick, and J.~Malik, ``Simultaneous
  detection and segmentation,'' in \emph{ECCV}, 2014.

\bibitem{mostajabi2014feedforward}
M.~Mostajabi, P.~Yadollahpour, and G.~Shakhnarovich, ``Feedforward semantic
  segmentation with zoom-out features,'' in \emph{CVPR}, 2015.

\bibitem{dai2014convolutional}
J.~Dai, K.~He, and J.~Sun, ``Convolutional feature masking for joint object and
  stuff segmentation,'' \emph{arXiv:1412.1283}, 2014.

\bibitem{eigen2014predicting}
D.~Eigen and R.~Fergus, ``Predicting depth, surface normals and semantic labels
  with a common multi-scale convolutional architecture,''
  \emph{arXiv:1411.4734}, 2014.

\bibitem{cogswell2014combining}
M.~Cogswell, X.~Lin, S.~Purushwalkam, and D.~Batra, ``Combining the best of
  graphical models and convnets for semantic segmentation,''
  \emph{arXiv:1412.4313}, 2014.

\bibitem{geiger1991parallel}
D.~Geiger and F.~Girosi, ``Parallel and deterministic algorithms from mrfs:
  Surface reconstruction,'' \emph{PAMI}, vol.~13, no.~5, pp. 401--412, 1991.

\bibitem{geiger1991common}
D.~Geiger and A.~Yuille, ``A common framework for image segmentation,''
  \emph{IJCV}, vol.~6, no.~3, pp. 227--243, 1991.

\bibitem{kokkinos2008computational}
I.~Kokkinos, R.~Deriche, O.~Faugeras, and P.~Maragos, ``Computational analysis
  and learning for a biologically motivated model of boundary detection,''
  \emph{Neurocomputing}, vol.~71, no.~10, pp. 1798--1812, 2008.

\bibitem{bell2014material}
S.~Bell, P.~Upchurch, N.~Snavely, and K.~Bala, ``Material recognition in the
  wild with the materials in context database,'' \emph{arXiv:1412.0623}, 2014.

\bibitem{papandreou2015weakly}
G.~Papandreou, L.-C. Chen, K.~Murphy, and A.~L. Yuille, ``Weakly- and
  semi-supervised learning of a dcnn for semantic image segmentation,'' in
  \emph{ICCV}, 2015.

\bibitem{zheng2015conditional}
S.~Zheng, S.~Jayasumana, B.~Romera-Paredes, V.~Vineet, Z.~Su, D.~Du, C.~Huang,
  and P.~Torr, ``Conditional random fields as recurrent neural networks,'' in
  \emph{ICCV}, 2015.

\bibitem{dai2015boxsup}
J.~Dai, K.~He, and J.~Sun, ``Boxsup: Exploiting bounding boxes to supervise
  convolutional networks for semantic segmentation,'' in \emph{ICCV}, 2015.

\bibitem{noh2015learning}
H.~Noh, S.~Hong, and B.~Han, ``Learning deconvolution network for semantic
  segmentation,'' in \emph{ICCV}, 2015.

\bibitem{liu2015semantic}
Z.~Liu, X.~Li, P.~Luo, C.~C. Loy, and X.~Tang, ``Semantic image segmentation
  via deep parsing network,'' in \emph{ICCV}, 2015.

\bibitem{chen2015semantic}
L.-C. Chen, J.~T. Barron, G.~Papandreou, K.~Murphy, and A.~L. Yuille,
  ``Semantic image segmentation with task-specific edge detection using cnns
  and a discriminatively trained domain transform,'' in \emph{CVPR}, 2016.

\bibitem{chen2014learning}
L.-C. Chen, A.~Schwing, A.~Yuille, and R.~Urtasun, ``Learning deep structured
  models,'' in \emph{ICML}, 2015.

\bibitem{schwing2015fully}
A.~G. Schwing and R.~Urtasun, ``Fully connected deep structured networks,''
  \emph{arXiv:1503.02351}, 2015.

\bibitem{chandra2016fast}
S.~Chandra and I.~Kokkinos, ``Fast, exact and multi-scale inference for
  semantic image segmentation with deep {G}aussian {CRF}s,''
  \emph{arXiv:1603.08358}, 2016.

\bibitem{GastalOliveira2011DomainTransform}
E.~S.~L. Gastal and M.~M. Oliveira, ``Domain transform for edge-aware image and
  video processing,'' in \emph{SIGGRAPH}, 2011.

\bibitem{bertasius2015high}
G.~Bertasius, J.~Shi, and L.~Torresani, ``High-for-low and low-for-high:
  Efficient boundary detection from deep object features and its applications
  to high-level vision,'' in \emph{ICCV}, 2015.

\bibitem{pinheiro2014weakly}
P.~O. Pinheiro and R.~Collobert, ``Weakly supervised semantic segmentation with
  convolutional networks,'' \emph{arXiv:1411.6228}, 2014.

\bibitem{pathakICCV15ccnn}
D.~Pathak, P.~Kr\"ahenb\"uhl, and T.~Darrell, ``Constrained convolutional
  neural networks for weakly supervised segmentation,'' 2015.

\bibitem{hong2015decoupled}
S.~Hong, H.~Noh, and B.~Han, ``Decoupled deep neural network for
  semi-supervised semantic segmentation,'' in \emph{NIPS}, 2015.

\bibitem{vezhnevets2011weakly}
A.~Vezhnevets, V.~Ferrari, and J.~M. Buhmann, ``Weakly supervised semantic
  segmentation with a multi-image model,'' in \emph{ICCV}, 2011.

\bibitem{liang2015proposal}
X.~Liang, Y.~Wei, X.~Shen, J.~Yang, L.~Lin, and S.~Yan, ``Proposal-free network
  for instance-level object segmentation,'' \emph{arXiv preprint
  arXiv:1509.02636}, 2015.

\bibitem{fowler2005redundant}
J.~E. Fowler, ``The redundant discrete wavelet transform and additive noise,''
  \emph{IEEE Signal Processing Letters}, vol.~12, no.~9, pp. 629--632, 2005.

\bibitem{vaidyanathan1990multirate}
P.~P. Vaidyanathan, ``Multirate digital filters, filter banks, polyphase
  networks, and applications: a tutorial,'' \emph{Proceedings of the IEEE},
  vol.~78, no.~1, pp. 56--93, 1990.

\bibitem{yu2015multi}
F.~Yu and V.~Koltun, ``Multi-scale context aggregation by dilated
  convolutions,'' in \emph{ICLR}, 2016.

\bibitem{dai2016rfcn}
J.~Dai, Y.~Li, K.~He, and J.~Sun, ``R-fcn: Object detection via region-based
  fully convolutional networks,'' \emph{arXiv:1605.06409}, 2016.

\bibitem{dai2016instance}
J.~Dai, K.~He, Y.~Li, S.~Ren, and J.~Sun, ``Instance-sensitive fully
  convolutional networks,'' \emph{arXiv:1603.08678}, 2016.

\bibitem{chen2015abc}
K.~Chen, J.~Wang, L.-C. Chen, H.~Gao, W.~Xu, and R.~Nevatia, ``Abc-cnn: An
  attention based convolutional neural network for visual question answering,''
  \emph{arXiv:1511.05960}, 2015.

\bibitem{sevilla2016optical}
L.~Sevilla-Lara, D.~Sun, V.~Jampani, and M.~J. Black, ``Optical flow with
  semantic segmentation and localized layers,'' \emph{arXiv:1603.03911}, 2016.

\bibitem{wu2016high}
Z.~Wu, C.~Shen, and A.~van~den Hengel, ``High-performance semantic segmentation
  using very deep fully convolutional networks,'' \emph{arXiv:1604.04339},
  2016.

\bibitem{shensa1992discrete}
M.~J. Shensa, ``The discrete wavelet transform: wedding the a trous and mallat
  algorithms,'' \emph{Signal Processing, IEEE Transactions on}, vol.~40,
  no.~10, pp. 2464--2482, 1992.

\bibitem{abadi2016tensorflow}
M.~Abadi, A.~Agarwal \emph{et~al.}, ``Tensorflow: Large-scale machine learning
  on heterogeneous distributed systems,'' \emph{arXiv:1603.04467}, 2016.

\bibitem{adams2010fast}
A.~Adams, J.~Baek, and M.~A. Davis, ``Fast high-dimensional filtering using the
  permutohedral lattice,'' in \emph{Eurographics}, 2010.

\bibitem{hariharan2011semantic}
B.~Hariharan, P.~Arbel{\'a}ez, L.~Bourdev, S.~Maji, and J.~Malik, ``Semantic
  contours from inverse detectors,'' in \emph{ICCV}, 2011.

\bibitem{liu2015parsenet}
W.~Liu, A.~Rabinovich, and A.~C. Berg, ``Parsenet: Looking wider to see
  better,'' \emph{arXiv:1506.04579}, 2015.

\bibitem{lin2014microsoft}
T.-Y. Lin \emph{et~al.}, ``Microsoft {COCO}: Common objects in context,'' in
  \emph{ECCV}, 2014.

\bibitem{Vemulapalli2016Gaussian}
R.~Vemulapalli, O.~Tuzel, M.-Y. Liu, and R.~Chellappa, ``Gaussian conditional
  random field network for semantic segmentation,'' in \emph{CVPR}, 2016.

\bibitem{yan2016combining}
Z.~Yan, H.~Zhang, Y.~Jia, T.~Breuel, and Y.~Yu, ``Combining the best of
  convolutional layers and recurrent layers: A hybrid network for semantic
  segmentation,'' \emph{arXiv:1603.04871}, 2016.

\bibitem{ghiasi2016laplacian}
G.~Ghiasi and C.~C. Fowlkes, ``Laplacian reconstruction and refinement for
  semantic segmentation,'' \emph{arXiv:1605.02264}, 2016.

\bibitem{arnab2015higher}
A.~Arnab, S.~Jayasumana, S.~Zheng, and P.~Torr, ``Higher order potentials in
  end-to-end trainable conditional random fields,'' \emph{arXiv:1511.08119},
  2015.

\bibitem{Shen2016Fast}
F.~Shen and G.~Zeng, ``Fast semantic image segmentation with high order context
  and guided filtering,'' \emph{arXiv:1605.04068}, 2016.

\bibitem{wu2016bridging}
Z.~Wu, C.~Shen, and A.~van~den Hengel, ``Bridging category-level and
  instance-level semantic image segmentation,'' \emph{arXiv:1605.06885}, 2016.

\bibitem{he2016identity}
K.~He, X.~Zhang, S.~Ren, and J.~Sun, ``Identity mappings in deep residual
  networks,'' \emph{arXiv:1603.05027}, 2016.

\bibitem{xia2015zoom}
F.~Xia, P.~Wang, L.-C. Chen, and A.~L. Yuille, ``Zoom better to see clearer:
  Huamn part segmentation with auto zoom net,'' \emph{arXiv:1511.06881}, 2015.

\bibitem{liang2015semantic}
X.~Liang, X.~Shen, D.~Xiang, J.~Feng, L.~Lin, and S.~Yan, ``Semantic object
  parsing with local-global long short-term memory,'' \emph{arXiv:1511.04510},
  2015.

\bibitem{liang2016semantic}
X.~Liang, X.~Shen, J.~Feng, L.~Lin, and S.~Yan, ``Semantic object parsing with
  graph lstm,'' \emph{arXiv:1603.07063}, 2016.

\bibitem{wang2014semantic}
J.~Wang and A.~Yuille, ``Semantic part segmentation using compositional model
  combining shape and appearance,'' in \emph{CVPR}, 2015.

\bibitem{wang2015joint}
P.~Wang, X.~Shen, Z.~Lin, S.~Cohen, B.~Price, and A.~Yuille, ``Joint object and
  part segmentation using deep learned potentials,'' in \emph{ICCV}, 2015.

\bibitem{badrinarayanan2015segnet}
V.~Badrinarayanan, A.~Kendall, and R.~Cipolla, ``Segnet: A deep convolutional
  encoder-decoder architecture for image segmentation,''
  \emph{arXiv:1511.00561}, 2015.

\bibitem{uhrig2016pixel}
J.~Uhrig, M.~Cordts, U.~Franke, and T.~Brox, ``Pixel-level encoding and depth
  layering for instance-level semantic labeling,'' \emph{arXiv:1604.05096},
  2016.

\bibitem{ronneberger2015u}
O.~Ronneberger, P.~Fischer, and T.~Brox, ``U-net: Convolutional networks for
  biomedical image segmentation,'' in \emph{MICCAI}, 2015.

\end{thebibliography}
}

%

\begin{IEEEbiography}[{\includegraphics[width=1in,height=1.25in,keepaspectratio]{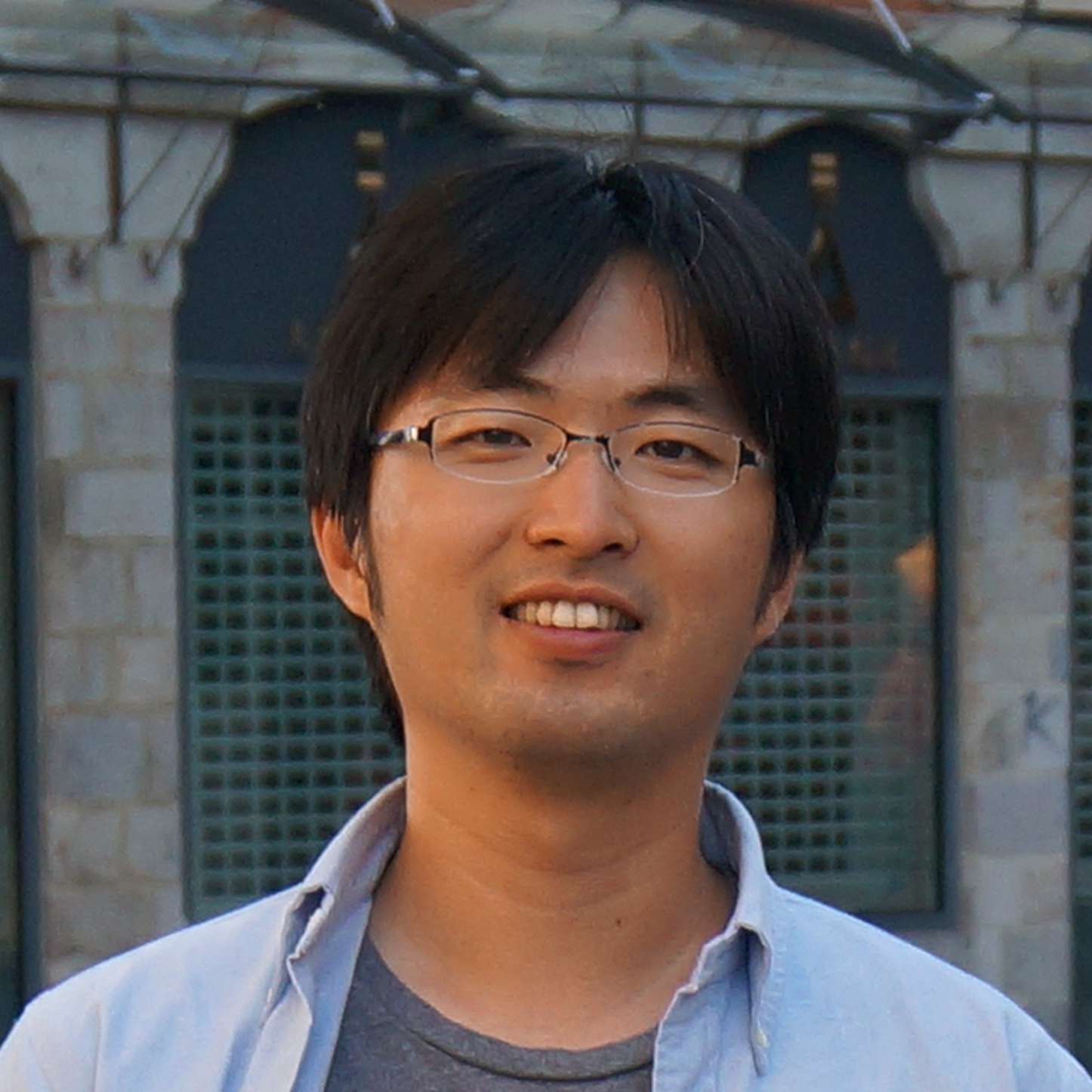}}]{Liang-Chieh Chen}
received his B.Sc. from National Chiao Tung University, Taiwan, his M.S. from the University of Michigan- Ann Arbor, and his
Ph.D. from the University of California- Los Angeles. He is currently working at Google. His research interests
include semantic image segmentation, probabilistic graphical models, and machine learning.
\end{IEEEbiography}

\begin{IEEEbiography}[{\includegraphics[width=1in,height=1.25in,keepaspectratio]{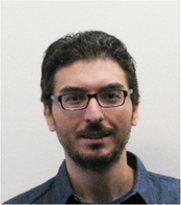}}]{George Papandreou}
(S'03--M'09--SM'14) holds a Diploma (2003) and a Ph.D. (2009) in Electrical
Engineering and Computer Science, both from the National Technical University
of Athens (NTUA), Greece. He is currently a Research Scientist at Google,
following appointments as Research Assistant Professor at the Toyota
Technological Institute at Chicago (2013-2014) and Postdoctoral Research Scholar
at the University of California, Los Angeles (2009-2013).

His research interests are in computer vision and machine learning, with a
current emphasis on deep learning. He regularly serves as a reviewer and
program committee member to the main journals and conferences in computer
vision, image processing, and machine learning. He has been a co-organizer of
the NIPS 2012, 2013, and 2014 Workshops on Perturbations, Optimization, and
Statistics and co-editor of a book on the same topic (MIT Press, 2016).
\end{IEEEbiography}

\begin{IEEEbiography}[{\includegraphics[width=1in,height=1.25in,keepaspectratio]{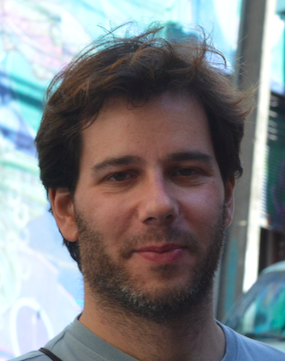}}]{Iasonas Kokkinos}
(S'02--M'06) obtained the Diploma of Engineering in 2001 and the Ph.D. Degree in 2006 from the School of Electrical and Computer Engineering of the National Technical University of Athens in Greece, and the Habilitation Degree in 2013 from Université Paris-Est. In 2006 he joined the University of California at Los Angeles as a postdoctoral scholar, and in 2008 joined as faculty the Department of Applied Mathematics of Ecole Centrale Paris (CentraleSupelec), working an associate professor in the Center for Visual Computing of CentraleSupelec and affiliate researcher at INRIA-Saclay. In 2016 he joined University College London and Facebook Artificial Intelligence Research. His currently research activity is on deep learning for computer vision, focusing in particular on structured prediction for deep learning, shape modeling, and multi-task learning architectures. He has been awarded a young researcher grant by the French National Research Agency, has served as associate editor for the Image and Vision Computing and Computer Vision and Image Understanding Journals, serves regularly as a reviewer and area chair for all major computer vision conferences and journals.
\end{IEEEbiography}

\begin{IEEEbiography}[{\includegraphics[width=1in,height=1.25in,keepaspectratio]{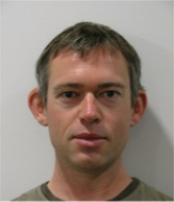}}]{Kevin Murphy}
was born in Ireland, grew
up  in  England,  went  to  graduate  school  in  the
USA (MEng from U. Penn, PhD from UC Berkeley,
Postdoc at MIT), and then became a professor at
the Computer Science and Statistics Departments at
the University of British Columbia in
Vancouver, Canada in 2004. After getting tenure,
Kevin  went  to  Google  in  Mountain  View,  California
for his sabbatical. In 2011, he converted
to a full-time research scientist at Google. Kevin
has published over 50 papers in refereed conferences
and journals related to machine learning and graphical models.
He  has  recently  published  an  1100-page  textbook  called
``Machine Learning: a Probabilistic Perspective''
(MIT Press, 2012).
\end{IEEEbiography}

\begin{IEEEbiography}[{\includegraphics[width=1in,height=1.25in,keepaspectratio]{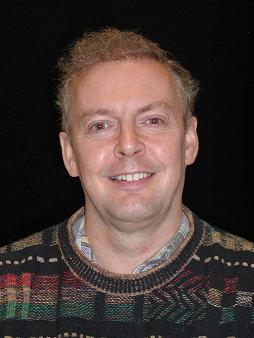}}]{Alan L. Yuille}
(F'09) received the BA degree in math-
ematics from the University of Cambridge in
1976. His PhD on theoretical physics, super-
vised by Prof. S.W. Hawking, was approved
in 1981. He was a research scientist in the
Artificial Intelligence Laboratory at MIT and
the Division of Applied Sciences at Harvard
University  from  1982  to  1988.  He  served
as  an  assistant  and  associate  professor  at
Harvard  until  1996.  He  was  a  senior  research scientist at the Smith-Kettlewell Eye
Research Institute from 1996 to 2002. He joined the University of
California, Los Angeles, as a full professor with a joint appointment
in  statistics  and  psychology  in  2002, and computer  science  in  2007. 
He was appointed a Bloomberg Distinguished Professor at Johns Hopkins University in January 2016.
He holds a joint appointment between the Departments of Cognitive science and Computer Science. 
His  research  interests  include
computational models of vision, mathematical models of cognition,
and artificial intelligence and neural network
\end{IEEEbiography}








\end{document}